\documentclass[journal]{IEEEtran}
\usepackage{amsmath,amsfonts}
\usepackage{algorithmic}
\usepackage{array}
\usepackage{textcomp}
\usepackage{stfloats}
\usepackage{url}
\usepackage{verbatim}
\usepackage{graphicx}
\usepackage[tight]{subfigure}
\def\BibTeX{{\rm B\kern-.05em{\sc i\kern-.025em b}\kern-.08em
    T\kern-.1667em\lower.7ex\hbox{E}\kern-.125emX}}
\usepackage{balance}

\usepackage{diagbox}
\usepackage{algorithm}
\usepackage{algorithmic}
\usepackage{mathtools}
\usepackage{bm}
\usepackage{soul,color, xcolor}
\soulregister\cite7 
\soulregister\citep7 
\soulregister\citet7 
\soulregister\ref7 
\soulregister\pageref7 
\renewcommand\hl[1]{#1}
\usepackage{booktabs}
\usepackage{bbding}
\usepackage{subcaption}
\usepackage{makecell,multirow}
\usepackage{colortbl}
\definecolor{mygray}{gray}{.9}
\definecolor{mypink}{rgb}{.99,.91,.95}
\definecolor{mycyan}{cmyk}{.3,0,0,0}
\usepackage[pagebackref=true,breaklinks=true, colorlinks,bookmarks=false]{hyperref}
\UseRawInputEncoding
\newcommand{\fakeparagraph}[1]{\vspace{1mm}\noindent\textbf{#1}}
\usepackage{cite}
\usepackage{pifont} 
\begin{document}
\title{Efficient and Robust Absolute Pose Estimation via Gravity-Prior-Driven Transformation Decoupling and Pose Refinement}
\author{Hu Cao\,$^{1\dagger}$, Qianyi Yang\,$^{2\dagger}$, Xinyi Li\,$^{2}$, Jiong Liu\,$^{2}$, Yinlong Liu\,$^{3*}$,  Alois Knoll\,$^{2}$,~\IEEEmembership{Fellow,~IEEE}
\thanks{$\dagger$ Equal contribution. $^*$Yinlong Liu is the corresponding author of this work (ylliu@cityu.edu.mo).}
\thanks{Authors Affiliation: $^{1}$School of Automation, Southeast University, Nanjing, China,
$^{2}$Chair of Robotics, Artificial Intelligence and Real-time Systems, Technical University of Munich, Munich, Germany,
		$^{3}$Faculty of Data Science, City University of Macau, Macau, China.}}

\markboth{Journal of \LaTeX\ Class Files,~Vol.~18, No.~9, September~2020}%
{How to Use the IEEEtran \LaTeX \ Templates}

\maketitle

\begin{abstract}
Estimation of the absolute pose of an object is an essential task for various robotic applications.  Recently, incorporating gravity direction as prior information has emerged as a popular approach to simplify absolute pose estimation. \hl{However, developing a robust and efficient algorithm to solve this challenging problem remains a difficult question due to large amounts of mismatches. In addition, obtaining an accurate pose solution from selected inlier correspondences with gravity prior is still a research gap.} In this paper, we propose a novel transformation strategy that exploits geometric relations derived from the gravity prior. Through transformation decoupling, the original 6 degrees of freedom (DoF) absolute pose estimation problem is simplified into a 4-DoFs problem: 1-DoF for the rotation angle and 3-DoFs for translation, significantly improving the efficiency. For the 1-DoF rotation angle, we apply a one-dimensional global voting algorithm for optimal estimation. Once the optimal rotation is obtained, the mismatched correspondences are preliminarily filtered, and  translation estimation, a linear problem, can be easily solved. Furthermore, to obtain accurate pose results, we introduce a novel pose refinement algorithm to enhance the accuracy of both rotation and translation. Extensive experiments on synthetic data and three publicly available real-world datasets (TUM RGB-D, ETH3D, and RobotCar) demonstrate that the proposed method achieves stronger performance compared to existing state-of-the-art (SOTA) approaches. To further validate our method, we integrated it into ORB-SLAM2. The results on the KITTI dataset show it effectively reduces drift and improves trajectory alignment during relocalization. The source code will be released upon acceptance.
\end{abstract}


\section{Introduction}
\IEEEPARstart{A}{bsolute} pose estimation is regarding calculating the camera pose from 2D image features and their 3D counterparts, which is a fundamental task for many applications of computer vision and robotics, such as augmented reality~\cite{marchand2015pose,sweeney2015efficient}, visual odometry~\cite{rehder2012global,li2020robust}, and visual localization~\cite{lynen2020large,jiao20202}.

\begin{figure}
\centering
\includegraphics[width=\linewidth]{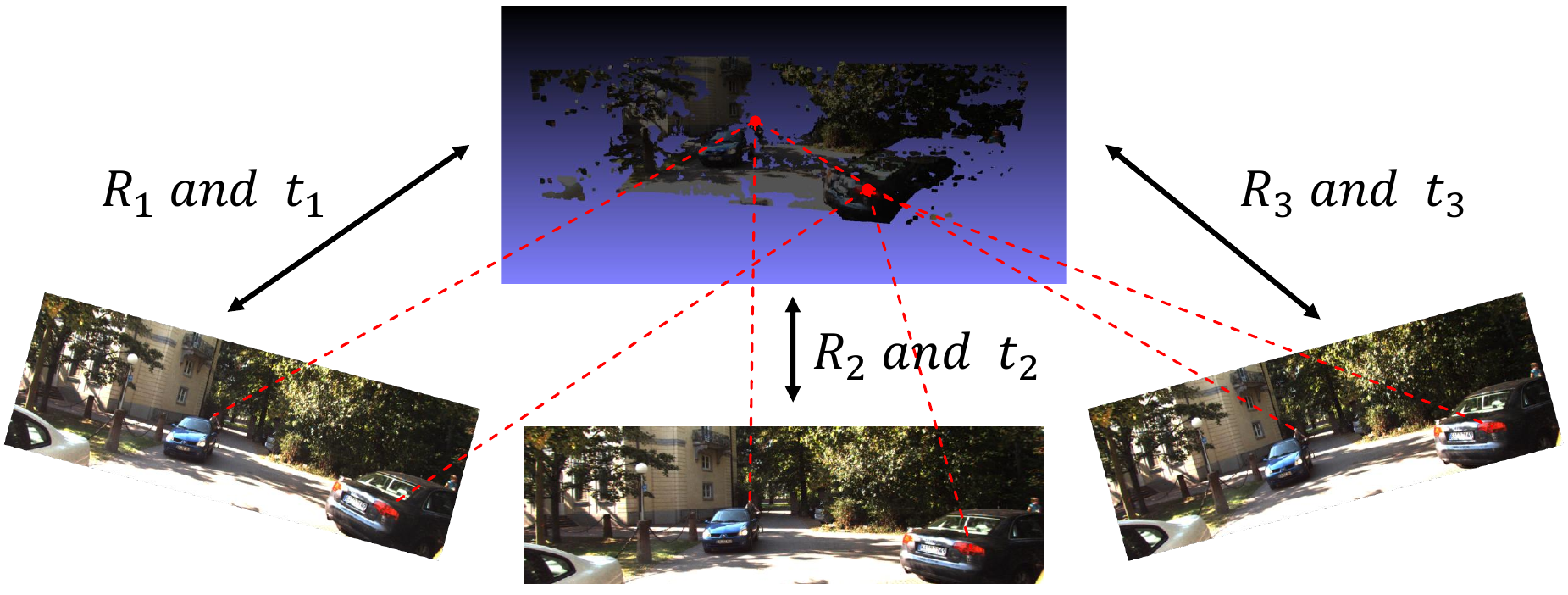}
\caption{The illustration of the absolute pose estimation problem. Technically, the objective is to estimate the rigid pose, comprising the rotation matrix $\mathbf{R}$ and the translation vector $\bm{t}$, from 2D-3D correspondences.}
\label{fig:intro}
\end{figure}

Technologically speaking, the task of estimating absolute pose involves determining the camera's orientation and position relative to a 3D reference world through the matching of 2D-3D features~\cite{kneip2014upnp,campbell2020globally} (see  Fig.~\ref{fig:intro}). It must be acknowledged that the pose can be estimated if the 2D and 3D correspondences are perfectly obtained, which is the well-known Perspective-n-Point (P$n$P) problem~\cite{garcia2024fast}.  However, it is almost impossible to have perfect 2D-3D feature matches in real applications~\cite{liu2023absolute}, and the input feature correspondences are typically contaminated by mismatches, which are also known as outliers~\cite{fischler1981random}. Unfortunately, precisely estimating the parameters from outlier-contaminated inputs is extremely hard (specifically, $\mathcal{NP}$-hard~\cite{Antonante22}), and even a small percentage of outliers might lead to wildly incorrect results~\cite{liu2022globally}.

Notably, in modern robotic sensor systems, inertial measurement units (IMUs) are typically integrated and are capable of providing prior knowledge of the gravity direction~\cite{geiger2013vision}. Understandably, the prior gravity can reduce the difficulty and help estimate the absolute pose of the camera~\cite{shao2021mofis,ding2021globally}. 
Additionally, the direction of gravity can be estimated by identifying vertical vanishing points or directions~\cite{salaun2016robust,almansa2003vanishing}, particularly in specific scenarios such as structured environments~\cite{micusik2013minimal}. Nonetheless, researchers utilize this prior directional information to enhance various tasks, including absolute pose estimation~\cite{lecrosnier2019camera,liu2023absolute}, relative pose estimation~\cite{ding2020homography,liu2021globally}, simultaneous localization and mapping (SLAM)~\cite{svarm2016city,ornhag2022trust}, and panoramic image stitching~\cite{ding2021minimal}. As a matter of fact, leveraging the gravity prior, certain parameters in camera orientation estimation can be simplified by applying geometric constraints~\cite{lecrosnier2019camera}. Technically, the full rigid pose estimation problem is reduced from 6-DoFs (three for rotation and three for translation) to 4-DoFs, as the rotation matrix is constrained to a single DoF for rotation about the specified axis~\cite{li2024transformation}.

Accordingly, we focus on 4-DoFs absolute pose estimation under the assumption that gravity directions are predetermined in this paper. Unfortunately, as far as we know, there is no effective and efficient way to remove outliers in advance~\cite{Antonante22}. Therefore, we target solving the
 absolute pose estimation with outlier-contaminated inputs. 


More specifically, as illustrated in Fig.~\ref{fig:pnp-problem}, the absolute pose estimation problem is commonly referred to as the well-known P$n$P problem~\cite{lepetit2009ep,gao2003complete}, and to estimate the absolute pose,  the 2D-3D feature correspondences should be constructed~\cite{campbell2020solving}. However, due to the limited performance (mismatches) of current feature matching methods~\cite{hadfield2018hard,zhou2018re} and challenges such as partial overlap and noisy data, the obtained 2D-3D correspondences often contain a significant proportion of outliers, which lead to serious incorrect results~\cite{zhou2018re}. In the practical applications, the de facto standard method to suppress the outlier is RANdom SAmple Consensus (RANSAC) method~\cite{fischler1981random}. However, as a non-deterministic algorithm, RANSAC only produces satisfactory results with a certain probability~\cite{le2019deterministic}. In other words, RANSAC method may provide an unsatisfactory pose solution occasionally~\cite{chin2017maximum}. Besides, its runtime grows exponentially with the outlier rate~\cite{bustos2017guaranteed}. 

For robotic applications, achieving an optimal solution with extreme robustness is essential, particularly in safety-critical scenarios~\cite{9546775,cheng2025urnet,10584449,antonante2021outlier}. To overcome this limitation, globally optimal and deterministic algorithms such as global voting~\cite{aiger2019general}, enumeration~\cite{svarm2016city}, and branch-and-bound (BnB)~\cite{hartley2009global,campbell2016gogma} have been employed. However, to search for the optimal solution, the time complexity of these highly robust algorithms increases substantially with the dimensionality of the optimization problem~\cite{liu2023absolute}. It should be mentioned that full rigid pose estimation, which is 6-DoFs, can not be considered a low-dimensional problem. In contrast, estimating the 4DOFs-gravity-known pose will be more efficient.

It is worth noting that to accelerate the globally optimal solution, transformation decoupling~\cite{jiao2021deterministic,liu2023absolute,li2023fast,li2024efficient,li2024transformation} has been recently proposed. Specifically, the strategy uses the geometric constraints of the rigid motion to decouple the original full estimation problem into sequential sub-problems, which can be solved more efficiently in lower-dimensional parameter spaces. Consequently, the final solution can be assembled by the optimal solutions to the sub-problems~\cite{li2024transformation}. This smart strategy has been verified to be effective in many pose estimation problems~\cite{li2024transformation}. \hl{
However, for the gravity-known pose estimation problem, the existing decoupling-based method~\cite{liu2023absolute} still has several shortcomings. Specifically, (1) it solves the problem by sequentially solving translation estimation (3-DoF) and rotation angle estimation (1-DoF). Unfortunately, estimating the 3-DoF translation is still not an easily solved low-dimensional problem, which retains high computational complexity. (2) In addition, the existing approach~\cite{liu2023absolute} only returns the inlier 2D-3D point correspondences, and it lacks a dedicated non-minimal refinement solution for the 4DoF gravity-constrained pose estimation problem, which still leaves a research gap~\cite{chandrasekhar2024posegravity}.  
}

Broadly speaking, existing gravity-aware 4-DoFs pose solvers have made important advances in reducing the dimensionality of the P$n$P problem, but two critical limitations remain unaddressed. First, mainstream methods adopt a translation-first decoupling strategy, which solves the 3-DoFs translation sub-problem with unfiltered outlier-contaminated correspondences, failing to fully exploit the low-dimensional advantage of the gravity-constrained rotation space and suffering from severe performance degradation under high outlier rates. Second, there is a lack of dedicated non-minimal refinement schemes tailored for the 4-DoFs gravity-constrained space, with most works relying on general P$n$P refinement that cannot fully leverage the gravity prior for further accuracy improvement. In this work, we address these limitations with a holistic rotation-first decoupling framework, consisting of three core original components: (1) a novel rotation-prioritized transformation decoupling strategy; (2) a 1D global voting scheme for robust rotation estimation and simultaneous outlier filtering; (3) a dedicated non-minimal refinement algorithm based on hidden variable resultant for high-precision 4-DoFs pose optimization. The following experiments systematically validate the independent contribution of each component, as well as the synergistic performance gain of our holistic framework.

In conclusion, our main contributions can be summarized as follows:
\begin{itemize}
    \item We establish a novel rotation-prioritized geometric relationship based on the gravity prior, which reduces the absolute pose estimation problem from 6-DoFs to 4-DoFs, and fundamentally improves the solving efficiency by reversing the conventional translation-first decoupling order.
    \item We design a 1D global voting algorithm tailored for the gravity-constrained 1DoF rotation space, which achieves deterministic, efficient rotation estimation and simultaneous outlier filtering, significantly enhancing the robustness of the method under high outlier rates.
    \item We propose a novel non-minimal refinement algorithm based on hidden variable resultant for the 4-DoFs gravity-aware pose estimation problem, which fills the research gap of high-precision optimization for this specific scenario and further improves the final pose accuracy.
    \item Extensive experiments on synthetic data, three real-world datasets, and a SLAM system integration validate that our method outperforms existing SOTA approaches, and can effectively reduce trajectory drift and improve relocalization performance in practical applications.
\end{itemize}




\section{Related Work}


\subsection{Outlier-free Absolute Pose Estimation}
Given 2D-3D correspondences without mismatches, numerous algorithms have been extensively studied~\cite{DBLP:journals/mta/MarulloTPV23}. The earliest known solution was introduced by Grunert~\cite{grunert1841pothenotische} in 1841, which addressed the minimal case of three projective point feature correspondences (P$3$P)~\cite{DBLP:conf/cvpr/0001YLOA23}. Since then, various closed-form solutions have been proposed, often formulating the problem as finding the roots of a polynomial equation~\cite{kneip2011novel,ke2017efficient}. For more than minimal inputs, i.e., P$n$P problem, a straightforward approach is to ignore the non-linear constraints and apply direct linear transformation (DLT) to estimate the absolute pose by solving a linear system~\cite{hartley2003multiple}. Alternatively, methods that fully consider non-linear constraints have been discussed in~\cite{kneip2014upnp}, where the core idea is to construct a system of polynomial equations derived from the original P$n$P problem and solve it using algebraic techniques~\cite{hartshorne2013algebraic}. In addition, it should be mentioned that Lepetit et al.~\cite{10.1007/s11263-008-0152-6} proposed the EPnP algorithm, which directly estimates the pose from four virtual control points~\cite{marchand2015pose}, offering a more efficient solution for the P$n$P problem. 

More recently, learning-based approaches have also been explored for absolute pose estimation~\cite{campbell2020solving,liu2022category,zhou2021semi}. Among these, transformer-based architectures have shown remarkable capability in modeling geometric dependencies. The Visual Geometry Grounded Transformer (VGGT)~\cite{wang2025vggt} embeds explicit geometric constraints into transformer layers for end-to-end pose learning and achieves SOTA accuracy on standard benchmarks via geometry-aware attention. In addition, RFMamba~\cite{zhang2025rfmamba} designs a frequency-aware state space model for RF-based perception, whose efficient frequency-domain noise processing provides insights for visual pose estimation with noisy correspondences. However, these learning-based methods~\cite{campbell2020solving,liu2022category,zhou2021semi,wang2025vggt,zhang2025rfmamba} require large-scale training data along with accurate ground truth poses to achieve good performance~\cite{zhou2025tumtraf,sarlin2021back,cao2024embracing}, which limits their deployment in scenarios where data annotation is scarce or expensive.

\subsection{Outlier-Robust Absolute Pose Estimation}
In practical applications, obtaining perfect 2D-3D correspondences is challenging due to many factors such as low image resolution and variations in point density. Consequently, many practical algorithms have been developed to suppress outlier correspondences and achieve robust pose estimation~\cite{fischler1981random}. Among these, RANSAC-based methods and their variants are widely employed when dealing with outlier-contaminated data~\cite{jiao20202}. However, these algorithms are inherently non-deterministic and cannot guarantee reliable results as the outlier rate increases~\cite{le2019deterministic}. To address this limitation, deterministic global optimization methods have been proposed~\cite{liu2023absolute}. Sv\"arm et al.~\cite{svarm2016city} introduced a deterministic approach for city-scale localization with a known vertical direction. Aiger et al.~\cite{aiger2019general} proposed a voting-based method to find the optimal absolute pose under the same prior. More recently, Liu et al.~\cite{liu2023absolute} employed a branch-and-bound (BnB) framework to determine the optimal translation, leveraging its strength in solving $\mathcal{NP}$-hard and non-convex problems~\cite{chin2017maximum}. As a matter of fact, even without 2D-3D correspondences, the BnB-based methods can still solve the absolute pose, although they will take a lot of time~\cite{campbell2020globally}. Despite their robustness, most deterministic methods are computationally intensive and too slow for real-time applications~\cite{liu2023absolute}.

\subsection{Transformation Decoupling}
In essence, the computational burden of robust pose estimation comes from the high dimensional space, which has to be searched globally, of the full rigid pose~\cite{chin2017maximum}.  One recently  popular way is decoupling the rotation and translation components and searching the separated spaces~\cite{chen2022deterministic,li2024transformation}. The key idea of the decoupling strategy is to explore the geometric constraints of the specific problem, and to construct sub-problems, which can be solved more efficiently. The most straightforward geometric constraints
are pairwise constraints. For example, the angle pairwise constraints are applied to solve 3D point cloud registration~\cite{Liu_2018_ECCV}. Similarly, the point-to-point pairwise constraints are also used to solve 3D point cloud registration~\cite{Yang9286491}.  Furthermore, the point-to-point pairwise constraints are used to solve the P$n$P problem~\cite{liu2019novel}. Besides, more complicated geometric constraints are explored in 3D point cloud registration~\cite{chen2022deterministic,li2024efficient,li2024transformation}, which significantly inspire our work.

\subsection{Known Gravity Direction}
With the widespread use of IMU sensors, the gravity direction can now be readily determined in advance, enabling numerous pose estimation techniques that leverage this prior information. Kukelova et al.~\cite{Kkelov2010ClosedFormST} proposed a closed-form solution for absolute camera pose estimation with a known gravity direction. Sweeney et al.~\cite{sweeney2015efficient} introduced another method that utilizes the gravity prior by projecting points in both the world and camera frames onto the gravity vector to form a constraint, ensuring that the distances between corresponding points remain consistent across both coordinate systems. Chandrasekhar~\cite{chandrasekhar2024posegravity} addressed the absolute pose estimation problem with an axis prior by enumerating the intersection of two conic curves.
Several other methods also exploit the known vertical direction in other pose estimation~\cite{li2024transformation}. For instance, the Perspective-n-Line (P$n$L) problem has been explored in~\cite{abdellali2018absolute,lecrosnier2019camera}, while relative pose estimation techniques with a gravity prior are discussed in~\cite{ding2021globally,ding2020homography}. The core idea behind these approaches is to reduce the problem’s dimensionality with the help of known gravity direction, thereby simplifying the computation.



\begin{figure}[!t]
\centering
\includegraphics[width=\linewidth]{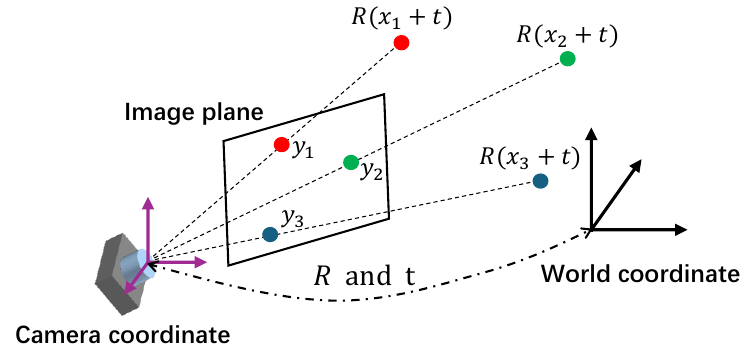}
\caption{The P$n$P problem involves determining the absolute camera pose from a set of given 2D-3D point pairs.}
\label{fig:pnp-problem}
\end{figure}

\section{Problem Formulation}
The Perspective-n-Point (P$n$P) problem is a fundamental geometric problem in computer vision. Its goal is to estimate the position and orientation (i.e., pose) of the calibrated camera coordinate system relative to the world coordinate system, given a set of 2D image points (pixel points) and their corresponding 3D world points. In other words, the P$n$P problem aims to find the camera’s external parameters (the rotation matrix $\mathbf{R}$ and translation vector $\bm{t}$).

With calibrated camera intrinsic parameters, the feature points' locations on the image can be normalized to the unit-sphere surface (i.e., $\mathbb{S}^2$) in the camera coordinate, where they are referred to as bearing vectors~\cite{campbell2020globally}. Specifically, for the $i$-th 3D observation $\bm{x}_i \in \mathbb{R}^3$ and its corresponding bearing vector $\bm{y}_i \in \mathbb{S}^2$, if they are perfectly aligned (see Fig.~\ref{fig:pnp-problem}),

\begin{equation}
	\lambda_i\bm{y}_i=\mathbf{R}\bm{x}_i+\bm{t},\quad i=1\cdots n
 \label{pair}
\end{equation}
where $\lambda_i$ is the $i$-th scale factor, and $n$ denotes the number of point pairs. The variables $\mathbf{R} \in \mathbb{SO}(3)$ and $\bm{t} \in \mathbb{R}^3$ represent the rigid motion to be determined. The P$n$P problem focuses on estimating the absolute camera pose, specifically $\mathbf{R}$ and $\bm{t}$, from a set of given 2D-3D point pairs.

\begin{figure*}[!t]
\centering
\includegraphics[width=\linewidth]{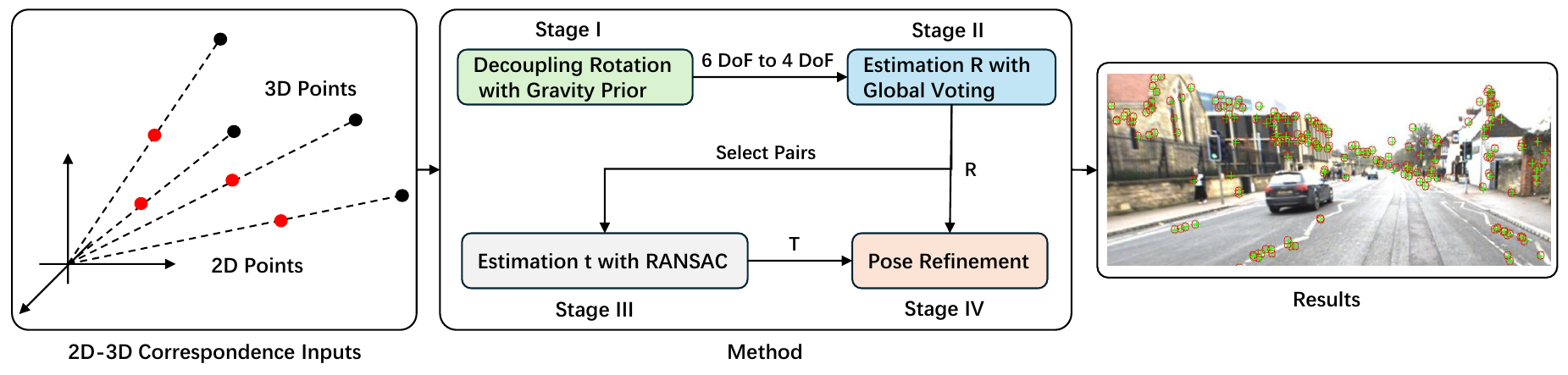}
\caption{The overview of our proposed absolute pose estimation method. It consists of four stages: decoupling rotation with gravity prior, estimation of rotation with global voting, estimation of translation with RANSAC from outlier-filtered inputs, and pose refinement.}
\label{fig:overview}
\end{figure*}

\section{Method}
\subsection{Overview}
As illustrated in Fig.~\ref{fig:overview}, our proposed absolute pose estimation method comprises four stages: decoupling rotation using a gravity prior, estimating rotation through global voting, estimating translation with RANSAC from outlier-filtered inputs, and pose refinement. Specifically, by leveraging the known gravity direction, the original P$n$P problem is reduced to a 4-DoFs problem: 1 DoF for the rotation angle and 3 DoFs for translation. For rotation angle estimation, we propose a novel geometric relation that transforms the problem into solving a corresponding trigonometric function. Global voting is then employed to identify the optimal rotation angle $\theta$. Once the optimal camera rotation $\mathbf{R}$ is determined, mismatched points are also significantly eliminated. The translation vector $\bm{t}$ in the 2D-3D correspondence problem is then efficiently computed using RANSAC. To further improve the accuracy of the rotation $\mathbf{R}$ and translation vector $\bm{t}$, pose refinement is performed. Notably, a hidden variable resultant is introduced to reduce computational complexity and ensure certifiable results.

\begin{figure}
    \centering
\includegraphics[width=\linewidth]{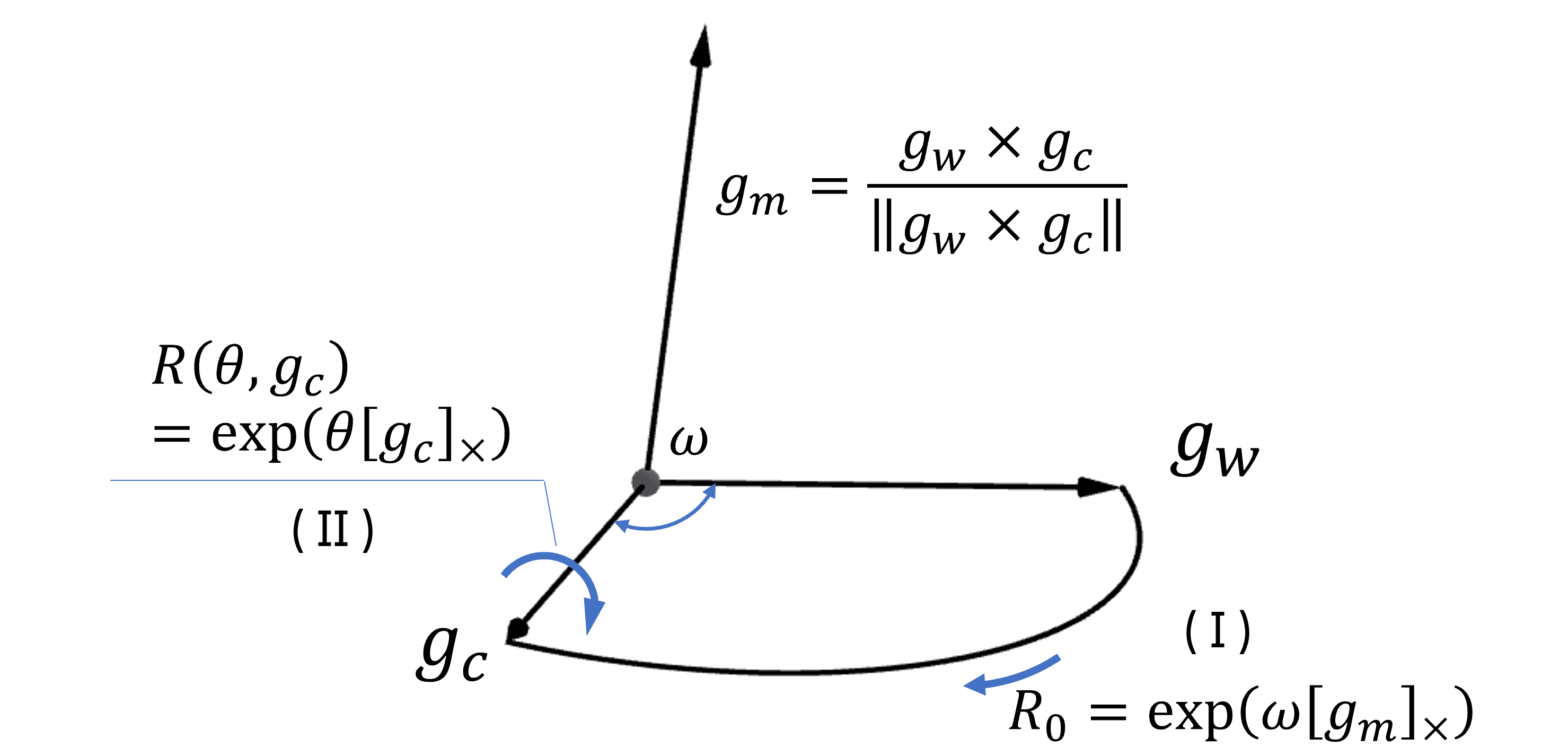}
    \caption{\hl{The rotation solution that is constrained by prior gravity, i.e., Eq.~(\ref{R_math}). Specifically, all the rotation solutions that meet the prior gravity constraint can be composed of two steps: (I) and (II). Please consult the main body of the text to understand the implications of all variables.}}
\label{fig:solution_of_Rot}
\end{figure}
\subsection{Decoupling Rotation with Gravity Prior} 
Decoupling rotation using a gravity prior reduces the problem from 6-DoFs to 4-DoFs. By obtaining the gravity directions $\bm{g}_c$ in the camera coordinate system and $\bm{g}_w$ in the world coordinate system, an additional constraint on the rotational motion can be established,
\begin{equation}
\bm{g}_c=\mathbf{R}\bm{g}_w
\end{equation}
The solution for the desired rotation $\mathbf{R}$ can be calculated explicitly~\cite{bustos2017guaranteed}(see Fig.~\ref{fig:solution_of_Rot}),
\begin{align}
	\mathbf{R} = \mathbf{R}\left(\theta,\bm{g}_c\right)\mathbf{R}_0 
 \label{R_math}
\end{align}
where $\mathbf{R}(\theta, \bm{g}_c)$ represents a rotation around the axis $\bm{g}_c$ by an angle $\theta$. This rotation can be computed using Rodrigues’ rotation formula~\cite{hartley2003multiple,hartley2009global},
\begin{equation}
\begin{aligned}
    \mathbf{R}\left(\theta,\bm{g}_c\right) &= \mathrm{exp}\left(\theta\left[\bm{g}_c\right]_\times\right) \\
    &=\mathbf{I}+\sin(\theta)[\bm{g}_c]_\times+\left(1-\cos(\theta)\right)[\bm{g}_c]_\times^2
    \label{rotation}
\end{aligned}
\end{equation}
where $\left[\bm{g}_c\right]_\times$ denotes the skew-symmetric matrix used for the cross product. Formally,
\begin{equation}
	\left[\bm{g}_c\right]_\times = \left[ \begin{array}{c}
g_c(1)\\
g_c(2)\\
g_c(3)
\end{array}
\right ]_\times = \left[ \begin{array}{ccc}
0 & -g_c(3) & g_c(2)\\
g_c(3) & 0 & -g_c(1)\\
-g_c(2) & g_c(1) & 0
\end{array} 
\right ]
\end{equation}

Similarly, $\mathbf{R}_0$ is a rotation that can align $\bm{g}_w$ with $\bm{g}_c$. The easy way is via the minimal/maximal geodesic motion, and it can be computed as~\cite{liu2023absolute}:
\begin{equation}
\begin{aligned}
    \mathbf{R}_0 &= \mathrm{exp}\left(\omega\left[\bm{g}_m\right]_\times\right) \\
    &=\mathbf{I}+\sin(\omega)[\bm{g}_m]_\times+\left(1-\cos(\omega)\right)[\bm{g}_m]_\times^2
    \label{R0}
\end{aligned}
\end{equation}
where $\omega=\arccos(\bm{g}_c^T\bm{g}_w)$ and $\bm{g}_m=\frac{\bm{g}_w \times \bm{g}_c}{|| \bm{g}_w \times \bm{g}_c ||}$. By substituting the Eq.~\eqref{R_math} back into the Eq.~\eqref{pair}, the original problem can be reformulated as:
\begin{equation}
\lambda_i\bm{y}_i=\mathbf{R}\left(\theta,\bm{g}_c\right)\mathbf{R}_0 \bm{x}_i+\bm{t},\quad i=1\cdots n
 \label{pairV2}
\end{equation}

Using Eq.~\eqref{R0}, $\mathbf{R}_0$ can be computed efficiently, leaving only the rotation angle $\theta$ of $\mathbf{R}(\theta, \bm{g}_c)$ to be determined. As a result, with the known gravity direction, the P$n$P problem is reduced to a 4-DoFs problem: 1 DoF for the rotation angle $\theta \in [-\pi, \pi]$ and 3 DoFs for the translation $\bm{t} \in \mathbb{R}^3$.

\subsection{The Estimation of Rotation Angle}
By considering pairwise constraints (2D-3D pairs), we can derive a relational equation based on Eq.~\eqref{pair}. Specifically, given two 2D-3D correspondences, we have
\begin{align}
	\left\{
	\begin{aligned}
		\lambda_i\bm{y}_i&=\mathbf{R}\bm{x}_i+\bm{t}
		\\
		\lambda_{i+1}\bm{y}_{i+1}&=\mathbf{R}\bm{x}_{i+1}+\bm{t}
	\end{aligned}
	\right.
\label{EQ7}
\end{align}

Eliminating $\lambda_i$ and $\lambda_{i+1}$,
\begin{align}
	\left(\bm{y}_i\times\bm{y}_{i+1}\right)^T\mathbf{R}\left(\bm{x}_i-\bm{x}_{i+1}\right) = 0
	\label{eq:pnp sub}
\Leftrightarrow\bm{m}_k^T\mathbf{R}\bm{n}_k=0
\end{align}
where $\times$ denotes the cross product. The terms $\bm{m}_k$ and $\bm{n}_k$ correspond to $\left(\bm{y}_i\times\bm{y}_{i+1}\right)$ and $\left(\bm{x}_i-\bm{x}_{i+1}\right)$, respectively. Notably, only the rotation $\mathbf{R}$ needs to be determined in Eq.~\eqref{eq:pnp sub}. Furthermore, with known gravity directions in both the camera and world coordinate systems, only the rotation angle $\theta$ remains to be solved. This can be accomplished using Rodrigues' rotation formula~\cite{hartley2003multiple,hartley2009global}. According to Eq.~\eqref{R_math} and Eq.~\eqref{eq:pnp sub}, 
\begin{align}
\bm{m}_k^T\mathbf{R}\left(\theta,\bm{g}_c\right)\mathbf{R}_0\bm{n}_k &= 0 
\end{align}

The equation can be reformulated as follows, based on Eq.~\eqref{rotation}:
\begin{align}
\bm{m}_k^T\left(\mathbf{I}+\sin(\theta)[\bm{g}_c]_\times+\left(1-\cos(\theta)\right)[\bm{g}_c]_\times^2\right)\mathbf{R}_0\bm{n}_k &= 0
\end{align}
Then, it can be simplified to
\begin{equation}
\left\{
\begin{aligned}
    {A}_k\sin(\theta)+{B}_k\cos(\theta)+{C}_k=0 \\
    \sin^2{(\theta)}+\cos^2{(\theta)}=1
\end{aligned}\right.
\end{equation}


where
${A}_k=\bm{m}_k^T[\bm{g}_c]_\times \mathbf{R}_0\bm{n}_k$, ${B}_k=-\bm{m}_k^T[\bm{g}_c]_\times^2 \mathbf{R}_0\bm{n}_k$ and ${C}_k=\bm{m}_k^T\mathbf{R}_0\bm{n}_k(1+[\bm{g}_c]_\times^2)$. 
The rotation angle $\theta$ can be computed by solving the corresponding trigonometric function. 
Specifically, global voting is employed to determine the optimal rotation angle $\theta^*$. For $N$ correspondences, the calculation is performed $\frac{N(N-1)}{2}$ times. To identify the most likely correct $\theta^*$, the range $[-\pi, \pi]$ is divided into 360 intervals, and the occurrences of $\theta$ in each interval are counted. The optimal rotation angle $\theta^*$ can be selected within the bin with the largest votes. 
If $\bm{m}_k$ and $\bm{n}_k$ cannot produce a $\theta$ in the bin with the largest votes, there should be at least an outlier for the 2D-3D correspondences. In contrast, if they produce a satisfactory $\theta$, the two correspondences are likely to be inliers. However, due to noise and randomness, we cannot guarantee that all correspondences are destined to be correct. Therefore, we count the number of each 2D-3D correspondence that produces $\theta$ in the interval where $\theta^*$ is located. The more often they appear, the more likely they are inliers~\cite{wu2025pointtruss}.
According to the frequency of occurrence, we can select the  top-ranked the data as the inner 2D-3D correspondences according to the application scenarios~\cite{moratalla2025clireg,Yan_2025_ICCV}. Notably, although this method cannot guarantee that all the remaining data are inliers, it can filter out a large part of outliers, which facilitates the next step of calculation.



\subsection{Translation Estimation}
After obtaining the optimal camera rotation $\mathrm{R}$, solving the translation then becomes a linear model fitting problem~\cite{chin2017maximum}. Besides, many mismatches can be removed by rotational constraints. Therefore, solving the sequential translation becomes quite easy. In this paper, the RANSAC algorithm is applied to estimate translation vector $\bm{t}$. The task is to determine $\bm{t}$ such that the transformed 3D points, after applying $\mathrm{R}$, align closely with their corresponding 2D bearing vectors. The key steps of this calculation process are shown in Algorithm~\ref{algorithm: RANSAC}. 


\begin{algorithm}[!t]
	\renewcommand{\algorithmicrequire}{\textbf{Input:}}
	\renewcommand{\algorithmicensure}{\textbf{Output:}}
	\caption{RANSAC for Estimating Translation Vector \( \bm{t} \) (Given \( \mathbf{R} \))}
	\label{algorithm: RANSAC}
	\begin{algorithmic}[1]
		\REQUIRE
		2D bearing vectors \( \bm{Y} = \{\bm{y}_1, \bm{y}_2, \dots, \bm{y}_i\} \), 3D world points \( \bm{X} = \{\bm{x}_1, \bm{x}_2, \dots, \bm{x}_i\} \), calculated rotation matrix \( \mathbf{R} \), 
        max iterations \( \bm{N}  \), error threshold \( \mathbf{threshold} \).
		\ENSURE
		Translation vector \( \bm{t_{\text{best}}} \).
        \STATE Initialize \( \bm{t_{\text{best}}} \gets \text{Null} \), \( \text{max\_inliers} \gets 0 \);

        \FOR{each iteration \( j = 0 \) to \( N - 1 \)}
            \STATE Randomly select two 2D-3D point pairs \( (\bm{x}_i, \bm{y}_i) \);
            \STATE Construct the overdetermined equation $\bm{Ax} = \bm{b}$ using Eq.~\eqref{EQ7};
            \STATE Compute translation vector $\bm{t}$ by solving an overdetermined equation using least squares;
            \STATE Initialize \( \text{inliers\_count} \gets 0 \);
            \FOR{each 2D \( \bm{y}_k \) and corresponding 3D \( \bm{x}_k \)}
                \STATE Compute bearing vector:
                    $\bm{y}_k^{\text{transformed}} \gets \mathbf{R} \cdot \bm{x}_k + \bm{t}$;
                \STATE Compute angle  between $\bm{y}_k$ and $\bm{y}_k^{\text{transformed}}$
                \IF{\( \text{angle} < \mathbf{threshold}  \)}
                    \STATE \( \text{inliers\_count} \gets \text{inliers\_count} + 1 \);
                \ENDIF
            \ENDFOR
            \IF{\( \text{inliers\_count} > \text{max\_inliers} \)}
                \STATE \( \text{max\_inliers} \gets \text{inliers\_count} \);
                \STATE \(  \bm{t_{\text{best}}} \gets \bm{t} \).
            \ENDIF
        \ENDFOR
        \RETURN \( \bm{t_{\text{best}}} \)
	\end{algorithmic}
\end{algorithm}

\subsection{Pose Refinement} 
To achieve more accurate estimates of the rotation $\mathbf{R}$ and the translation vector $\bm{t}$, pose refinement is employed in this work. The inliers derived from the initial estimates of $\mathbf{R}$ and $\bm{t}$ are used to refine and re-compute these values. Specifically, the pose estimation problem is formulated using linear projective constraints as follows:
\begin{equation}
	\left[\bm{y}_i\right]_\times(\mathbf{R}\bm{x}_i+\bm{t}) = 0,\quad i=1\cdots n
 \label{constraints}
\end{equation}

For a larger number of 2D-3D correspondences, the left and right sides of Eq.~\eqref{constraints} may not be exactly equal. In such cases, the goal is to estimate the optimal $\mathbf{R}$ and $\bm{t}$ in the least squares sense. Therefore, the optimizable objective function can be defined to minimize the errors as follows:

\begin{equation}
	\min_{\bm{R,t}}\sum_{i}(\left[\bm{y}_i\right]_\times(\mathbf{R}\bm{x}_i+\bm{t}))^T(\left[\bm{y}_i\right]_\times(\mathbf{R}\bm{x}_i+\bm{t}))
\label{ObjectFuncV1}
\end{equation}

\fakeparagraph{Constrained least squares.}  
Note that $\mathbf{R}\bm{x}$, the rotation of a vector $\bm{x}$, can be expressed as:

\begin{equation}
\left[ \begin{array}{ccccccccc}
x_1 & x_2 & x_3&0 & 0 & 0 & 0 & 0&0\\
0 & 0 & 0 & x_1 & x_2 & x_3& 0 & 0&0\\
0 & 0 & 0 & 0 & 0&0&x_1 & x_2 & x_3
\end{array} 
\right ] \mathrm{vec}(\mathbf{R})
\end{equation}
where $\mathrm{vec}(\mathbf{R})$ represents the row-flattened vector form of $\mathbf{R}$. In this work, we align $\bm{g}_c$ with the z-axis, the rotation matrix contains only three unique elements, allowing the expression to be simplified as follows:

\begin{equation}
	\mathbf{R}\bm{x} = \left[ \begin{array}{ccc}
\cos(\theta) & -\sin(\theta) & 0\\
\sin(\theta) & \cos(\theta) & 0\\
0 & 0 & 1
\end{array} 
\right ] \left[ \begin{array}{c}
x_1\\
x_2\\
x_3
\end{array}
\right ] 
\end{equation}
where $x=\cos(\theta)$ and $y=\sin(\theta)$.
It can be transformed into the following form:

\begin{equation}
	\mathbf{R}\bm{x} =\bm{Xr} = \left[ \begin{array}{ccc}
x_1 & -x_2& 0\\
x_2 & x_1 & 0\\
0 & 0 &x_3
\end{array} 
\right ] \left[ \begin{array}{c}
x\\
y\\
1
\end{array}
\right ] 
\end{equation}
where $\bm{X}$ represents the matrix form of the vector $\bm{x}$. Consequently, the optimizable objective function of Eq.~\eqref{ObjectFuncV1} can be rewritten as follows:
\begin{equation}
	\min_{\bm{r},\bm{t}}\sum_{i}(\bm{X}_i\bm{r}+\bm{t})^T\bm{Q}_i(\bm{X}_i\bm{r}+\bm{t})
\label{ObjectFunc}
\end{equation}
where $\bm{Q}_i$ represents $\left[\bm{y}_i\right]_\times^T\left[\bm{y}_i\right]_\times$. This sum of squared linear equations is non-negative and convex concerning the unconstrained $\bm{t}$. By taking the derivative of the objective function for $\bm{t}$ and setting it to zero, the globally minimizing value of $\bm{t}$ can be computed given $\bm{r}$:

\begin{equation}
\begin{aligned}
\bm{t} &= -\bm{Sr}\\
&=-(\sum_{i}\bm{Q}_i)^{-1} (\sum_{i}\bm{Q}_i\bm{X}_i)\bm{r}
\end{aligned}
\end{equation}

By substituting the value of $\bm{t}$ back into the objective function (Eq.~\ref{ObjectFunc}) and consolidating the terms, the simplified constrained quadratic minimization problem can be expressed as follows:

\begin{equation}
\begin{aligned}
&\min_{\bm{r},\bm{t}}\sum_{i}(\bm{X}_i\bm{r}-\bm{Sr})^T\bm{Q}_i(\bm{X}_i\bm{r}-\bm{Sr})\\
&=\min_{\bm{r}}\bm{r}^T\bm{\Omega}\bm{r} \quad s.t. \quad x^2+y^2 =1.\\
\end{aligned}
\label{Problem}
\end{equation}
where $\bm{\Omega}$ is represented as follows:
\begin{equation}
\bm{\Omega}=\sum_{i}(\bm{X}_i - \bm{S})^{T} \bm{Q}_i(\bm{X}_i - \bm{S}) \\
\end{equation}

\fakeparagraph{Solution with hidden variable resultant.} Clearly, Eq.~\eqref{Problem} represents a typical equality-constrained optimization problem~\cite{bertsekas2014constrained}. The corresponding Lagrangian formulation is:

\begin{equation}
    \min_{x,y}f(x,y)+\beta(x^2+y^2-1)
\end{equation}
where $\beta$ is the Lagrangian multiplier, and $f(x, y)$ represents the expanded polynomial form of $\bm{r}^T\bm{\Omega}\bm{r}$. By taking the partial derivatives with respect to $x$, $y$, and $\beta$:

\begin{equation}
    \frac{df}{dx}+2\beta x=0
\label{first_equation}
\end{equation}
\begin{equation}
    \frac{df}{dy}+2\beta y=0
\label{second_equation}
\end{equation}
\begin{equation}
    x^2+y^2=1
\label{circle}
\end{equation}

The term $\beta$ can be eliminated by multiplying Eq.~\eqref{first_equation} by $y$ and subtracting Eq.~\eqref{second_equation} multiplied by $x$:

\begin{equation}
    y\frac{df}{dx}-x\frac{df}{dy}=0
\label{newEq}
\end{equation}

More specifically, Eq.~\eqref{newEq} can  be expressed in a conic form, referred to as the derivative conic:

\begin{equation}
\begin{aligned}
  \begin{bmatrix}
x\\
y\\
1
\end{bmatrix}^T&\begin{bmatrix}
     -2\mathbf{\Omega_{0,1}}&\mathbf{\Omega_{0,0}}-\mathbf{\Omega_{1,1}}&-\mathbf{\Omega_{1,2}} \\
     \mathbf{\Omega_{0,0}}-\mathbf{\Omega_{1,1}}&2\mathbf{\Omega_{0,1}}&\mathbf{\Omega_{0,2}}\\
     -\mathbf{\Omega_{1,2}}&\mathbf{\Omega_{0,2}}&0
\end{bmatrix}\begin{bmatrix}
x\\
y\\
1
\end{bmatrix}\\
&=\bm{r}^T\bm{\Lambda}\bm{r} =0
\end{aligned}
\label{conic}
\end{equation}

This conic generally represents a hyperbola. The Eq.~\eqref{circle} defines the equation of a circle, which can also be expressed in conic form. Specifically, solving Eq.~\eqref{circle} and Eq.~\eqref{conic} can be framed as finding the intersection of two conic curves, a problem comprehensively discussed by Prof. Jürgen Richter-Gebert in~\cite{richter2011perspectives}. The core idea involves constructing a new degenerate conic, which can consist of two possibly coincident lines. The points of intersection between this degenerate conic and either of the original conics yield the desired points. In~\cite{chandrasekhar2024posegravity}, the authors addressed this problem using this geometric approach. However, a drawback of this method is the need to identify all possible intersections, which increases computational complexity and susceptibility to calculation errors. In this work, we introduce the use of a hidden variable resultant~\cite{hartley2012efficient} to address this problem. Specifically, from Eq.~\eqref{circle} and Eq.~\eqref{conic}, we derive:

\begin{equation}
\left\{
\begin{array}{cc}
     ax^2+by^2+cxy+dx+ey&=0  \\
     x^2+y^2-1 &=0 \\
     ax^3+bxy^2+
     cx^2y+dx^2+exy&=0\\
     x^3+xy^2-x&=0
\end{array} \right.
\end{equation}
where $a=\mathbf{\Omega}_{0,1}$, $b=-\mathbf{\Omega}_{0,1}$, $c=\mathbf{\Omega}_{1,1}-\mathbf{\Omega}_{0,0}$, $d=\mathbf{\Omega}_{1,2}$, $e=-\mathbf{\Omega}_{0,2}$. The equations can then be rewritten as follows:

\begin{equation}
  \begin{bmatrix}
     0&a&cy+d&by^2+ey  \\
     0&1&0&y^2-1 \\
     a&cy+d&by^2+ey&0\\
    1&0&y^2-1&0
\end{bmatrix}\begin{bmatrix}
x^3\\
x^2\\
x\\
1
\end{bmatrix}=\begin{bmatrix}
0\\
0\\
0\\
0
\end{bmatrix}
\end{equation}

To obtain the non-trivial solutions, the determinant of the coefficient matrix should be zero.
 \begin{equation}
 \centering
\begin{vmatrix}
     0&a&cy+d&by^2+ey  \\
     0&1&0&y^2-1 \\
     a&cy+d&by^2+ey&0\\
    1&0&y^2-1&0
\end{vmatrix} =0
\end{equation}
This leads to a univariate polynomial in terms of the hidden variable,

\begin{equation}
    m_4y^4+m_3y^3+m2y^2+m_1y+m_0=0
\end{equation}
where $m_4=(a-b)^2+c^2$, $m_3=2cd-2e(a-b)$, $m_2=-c^2+d^2+e^2-2a(a-b)$, $m_1=2ae-2cd$, $m_0=a^2-d^2$. By solving the univariate polynomial system, we can obtain $y$. Then, $x$ can be solved as follows:

\begin{equation}
   x=\pm \sqrt{1-y^2}
\end{equation}
Finally, we select the $(x, y)$ that minimizes $\mathbf{r}^T\mathbf{\Omega}\mathbf{r}$ as the result.
\section{Experimental Setting}

\subsection{Compared Methods}
In our experiments, we compare our proposed pose estimation method against several existing approaches on both synthetic and real-world datasets. Specifically, the following algorithms are included in the comparison:
 \begin{itemize}
    \item BnB: The 4-DoF algorithm based on the Branch-and-Bound framework proposed in~\cite{liu2023absolute}.
    \item Gao: The P3P algorithm proposed by Gao et al.~\cite{1217599}, embedded within RANSAC.
    \item Ap3: The AP3P algorithm introduced by Ke et al.~\cite{8099974}, incorporated into the RANSAC framework.
    \item Swee: The P2P algorithm proposed by Sweeney et al.~\cite{7328054}, encapsulated within the RANSAC framework.
    \item Kuke: The P2P algorithm introduced by Kukelova et al.~\cite{Kkelov2010ClosedFormST}, integrated with RANSAC.
    \item Sqpnp: The SqPnP algorithm proposed by Terzakis et al.~\cite{10.1007/978-3-030-58452-8_28}, embedded within RANSAC.
    \item Epnp: The EPnP algorithm proposed by Lepetit et al.~\cite{10.1007/s11263-008-0152-6}, integrated with RANSAC.
    \item \hl{SupeRANSAC: The PnP algorithm proposed by Barath~\cite{barath2025superansac}, which represents the SOTA in RANSAC-like methods.}
    \item \hl{AaPnP: The axis-aligned PnP algorithm proposed by Roch et al.~\cite{roch2024axes}, which is a method for camera pose estimation that exploits specific real-world constraints.}
\end{itemize}

\subsection{Evaluation Metrics}
To compare and evaluate the performance of each method, we employ two evaluation metrics: rotation error and translation error. The rotation error, measured by angle distance~\cite{hartley2013rotation}, is defined as follows:

\begin{equation}
    e_{rot} = \arccos\left(\frac{Tr(\mathbf{R}_{gt}^T\mathbf{R}_{cal})-1}{2}\right)
\end{equation}
where $\mathit{Tr(\cdot)}$ denotes the trace of a square matrix, and $\mathbf{R}_{cal}$ represents the rotation result obtained from the implemented algorithm. Furthermore, the translation error is defined as follows:
\begin{equation}
    e_{trans}=\left\| \bm{t}_{gt}-\bm{t}_{cal} \right\|
\end{equation}
where $\bm{t}_{cal}$ denotes the translation result obtained from the corresponding algorithms.

\subsection{Hardware Setting}
To ensure a fair comparison, all experiments are conducted under the same hardware settings, and all methods are implemented in C++, and the experiments are performed on a personal laptop equipped with an R7-6800H CPU @ 3.20 GHz and 16 GB of RAM.

\begin{figure*}[!t]
	\subfigure[]{
		\includegraphics[width=0.23\linewidth]{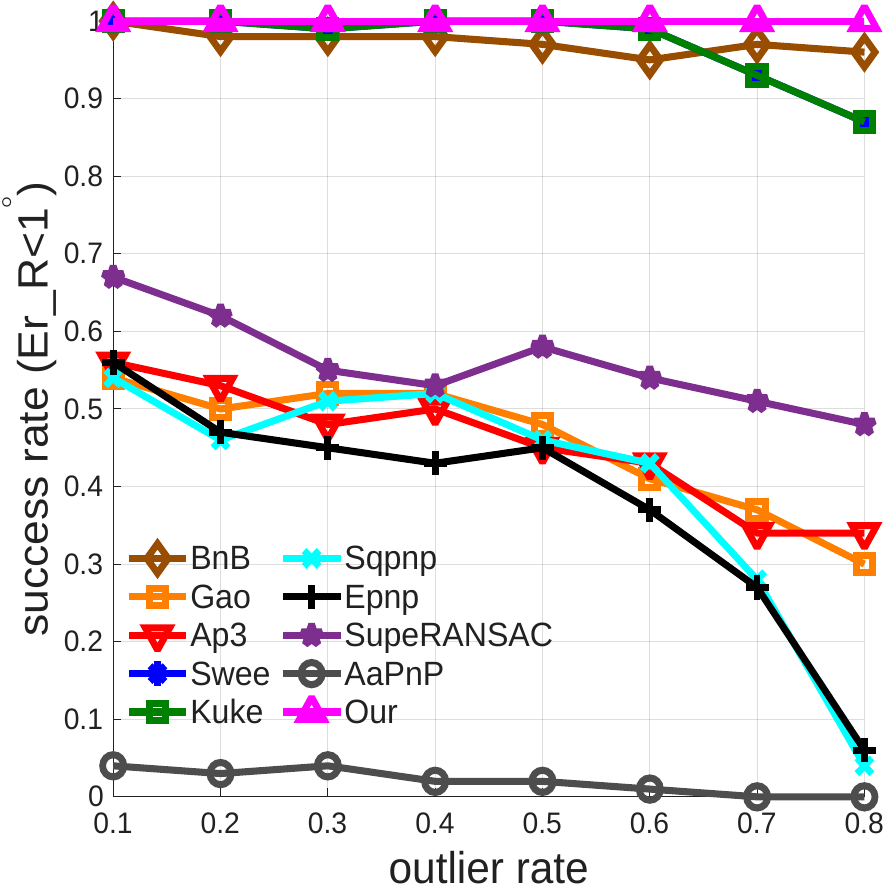}
		\label{fig:OrR}
	}
	\subfigure[]{
		\includegraphics[width=0.23\linewidth]{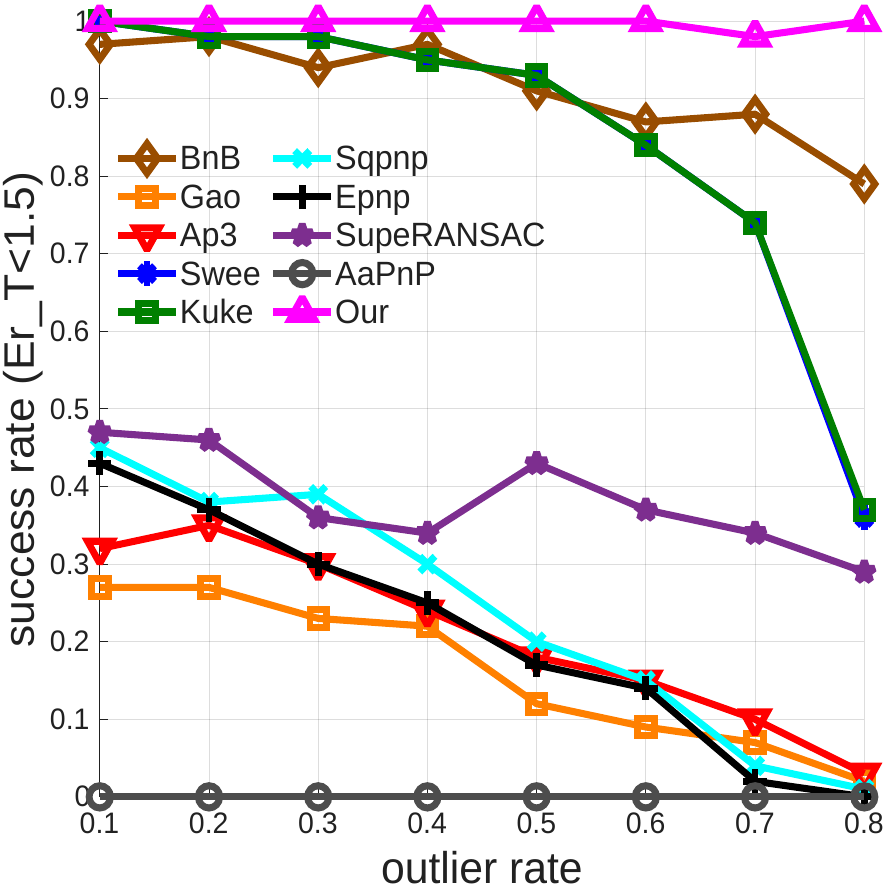}
		\label{fig:OrT}
	}
	\subfigure[]{
		\includegraphics[width=0.23\linewidth]{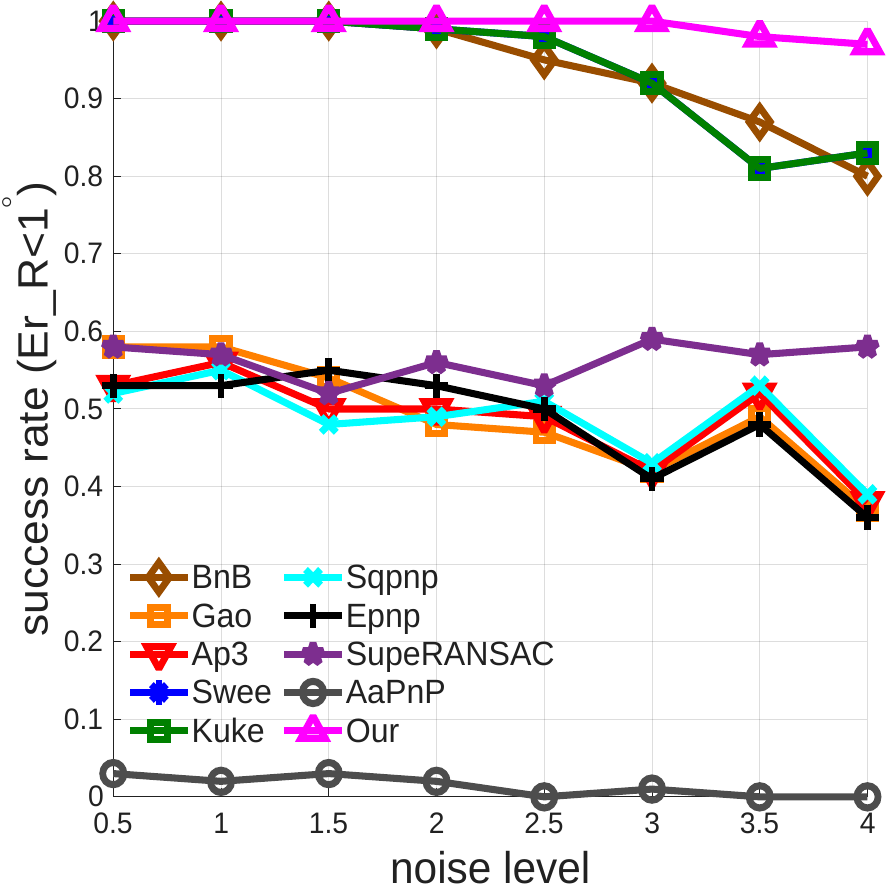}
		\label{fig:SigR}
	}
	\subfigure[]{
		\includegraphics[width=0.23\linewidth]{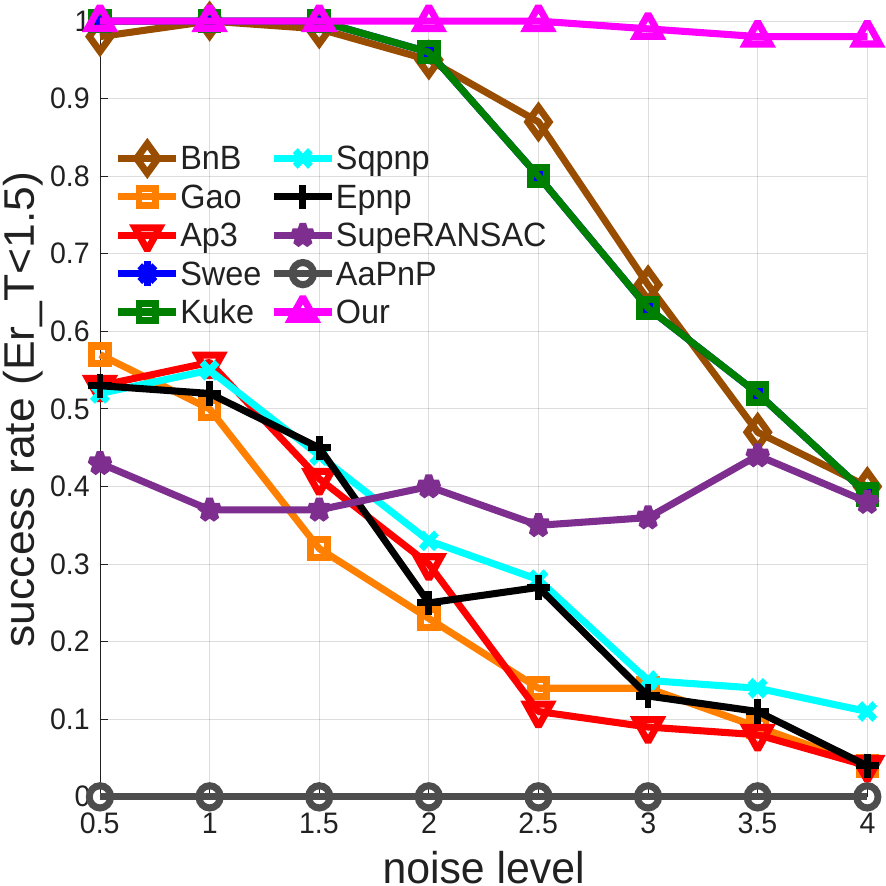}
		\label{fig:SigT}
	}
 \caption{\hl{Controlled Experiments. (a) and (b) illustrate the success rates for input data under varying outlier rates. (c) and (d) depict the success rates for input data with different noise levels.}}
 \label{fig:Or_Sig}
 \centering
\end{figure*}

\section{Experimental Results}

\subsection{Synthetic Data Experiments}
\fakeparagraph{Data generation.}
In our simulated experiments, we first randomly generate the pose of the truth of the ground, with the rotation matrix $\mathbf{R_{gt}} \in \mathbb{SO}(3)$ and the translation vector $\bm{t_{gt}} \in {[-100,100]}^3$. To obtain 2D-3D correspondences, 3D points are randomly generated within the range $[-100,100]^3$, and their corresponding bearing vectors are computed by applying a rigid transformation and normalizing them to unit length. Outliers are introduced by randomly generating new 3D points to replace the original ones, with the outlier rate denoted as $\omega = N_{outlier}/N$. To simulate noise, we apply a standard normal distribution with a deviation parameter $\sigma$. Noise can be added in two ways: either by perturbing the 3D points directly or by adding noise to pixel coordinates, which requires setting the focal length and intrinsic matrix.


\fakeparagraph{Controlled experiments.}
To evaluate the performance of different methods, we conduct controlled experiments on synthetic data, focusing on two key aspects: outlier rate and noise level. To clearly illustrate the results, we use the success rate as the evaluation metric. Specifically, Er\_R $< 1^\circ$ means a successful case for rotation calculation that $e_{rot}$ is less than 1 degree. Similarly, Er\_T $< 1.5$ represents a good case for translation calculation that $e_{trans}$ is less than 1.5.

First, we evaluate the robustness of each method against outliers. The input data are tested with different outlier rates, specifically $\omega = \{0.1,0.2,\cdots,0.8\}$. In these experiments, the number of 2D-3D correspondences is set to $N = 1000$, the noise level is fixed at $\sigma = 2.0$, and each scenario is repeated 100 times. The results are presented in Fig.~\ref{fig:OrR} and Fig.~\ref{fig:OrT}. It can be observed that, under the same noise level, our proposed method exhibits significantly higher robustness in both rotation and translation estimation as the outlier rate increases compared to the other methods.

Second, we evaluate the robustness against noise. The input data is tested under various noise levels, i.e., $\sigma = \{0.5,1,\cdots,4\}$. Specifically, we set the number of correspondences to $N = 1000$ and fix the outlier rate at $\omega = 0.5$. Each experiment is repeated 100 times. The results are shown in Fig.~\ref{fig:SigR} and Fig.~\ref{fig:SigT}. We observe that, compared to the outlier rate, the noise level has a more significant impact on the performance. However, within a reasonable noise range, our method maintains an almost 100\% success rate in both rotation and translation estimations.

\begin{figure}[!t]
	\subfigure[]{
		\includegraphics[width=0.45\linewidth]{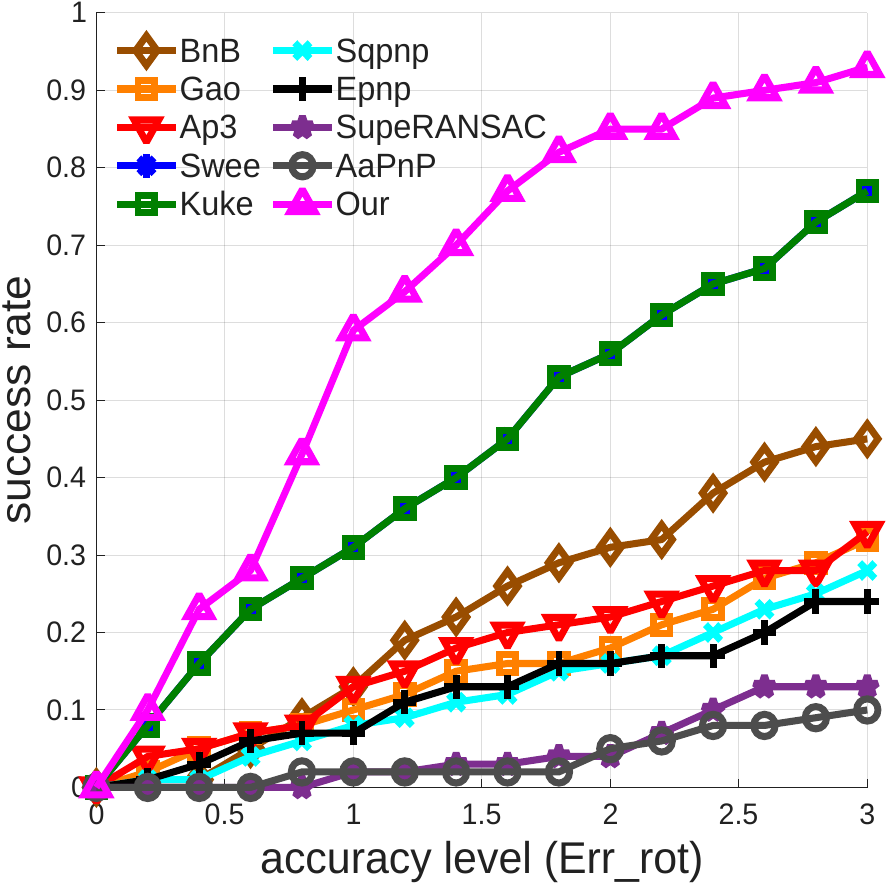}
		\label{fig:AcuR}
	}
	\subfigure[]{
		\includegraphics[width=0.45\linewidth]{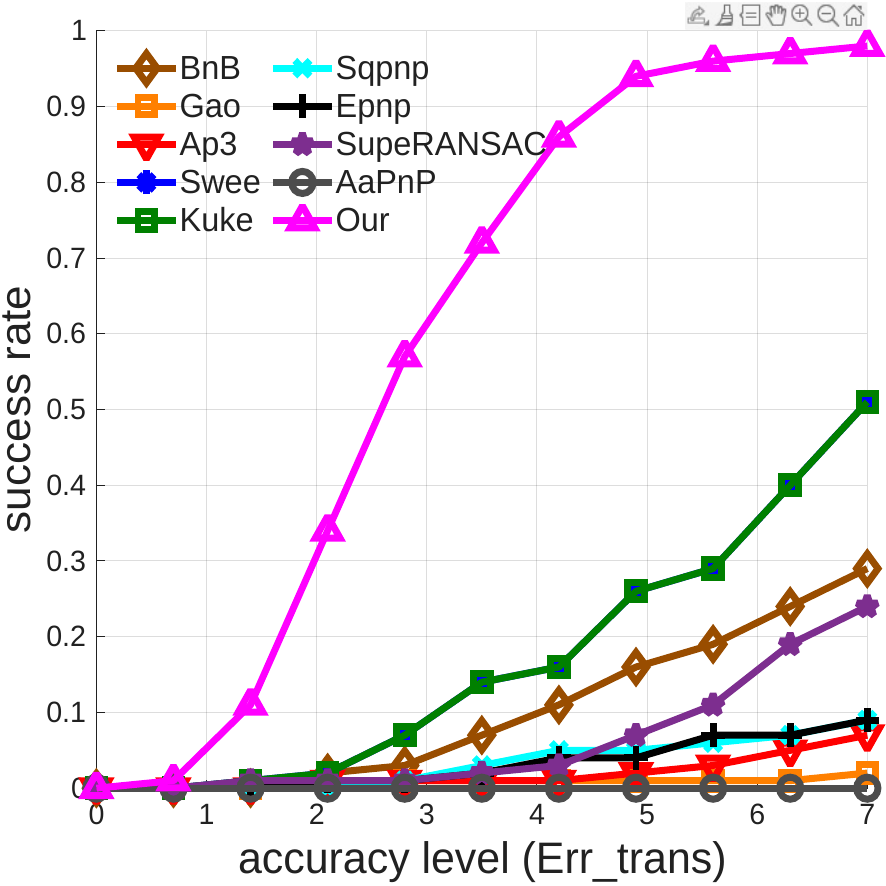}
		\label{fig:AcuT}
	}
 \caption{\hl{Accuracy level-success rate curves under heavily corrupted data conditions, showing the performance of each method in scenarios where both outlier rates and noise levels are high.}}
 \label{fig:Acu}
 \centering
\end{figure}

Finaly, to further demonstrate the robustness of the proposed method, we conduct experiments in extreme scenarios where the input data is highly corrupted with both a large outlier rate and a high noise level. Specifically, we set the number of correspondences to $N = 1000$, the outlier rate to $\omega = 0.5$, and the noise level to $\sigma = 10.0$. Each experiment is repeated 100 times. As illustrated in Fig.~\ref{fig:Acu}, our proposed method demonstrates significantly greater robustness to both noise and outliers compared to other methods, maintaining a higher success rate even under severe conditions.



\begin{figure}[!t]
	\subfigure[]{
		\includegraphics[width=0.45\linewidth]{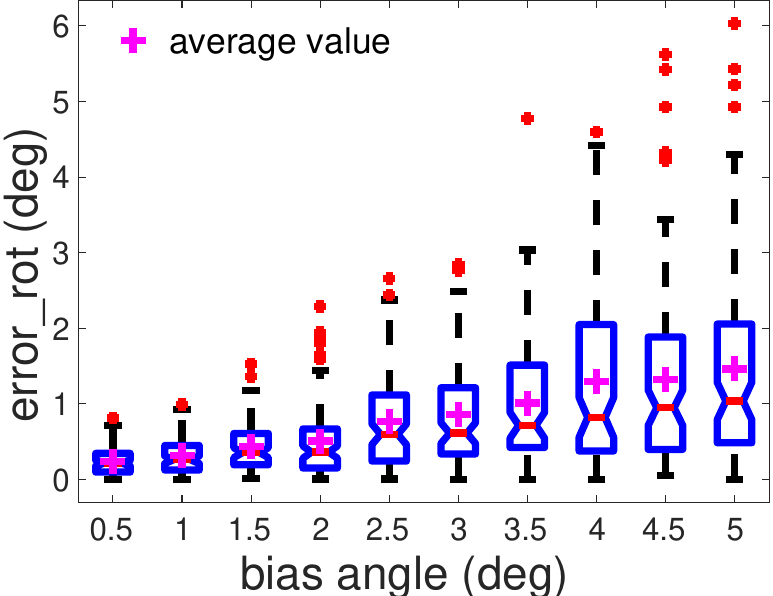}
		\label{fig:gbiasR}
	}
	\subfigure[]{
		\includegraphics[width=0.45\linewidth]{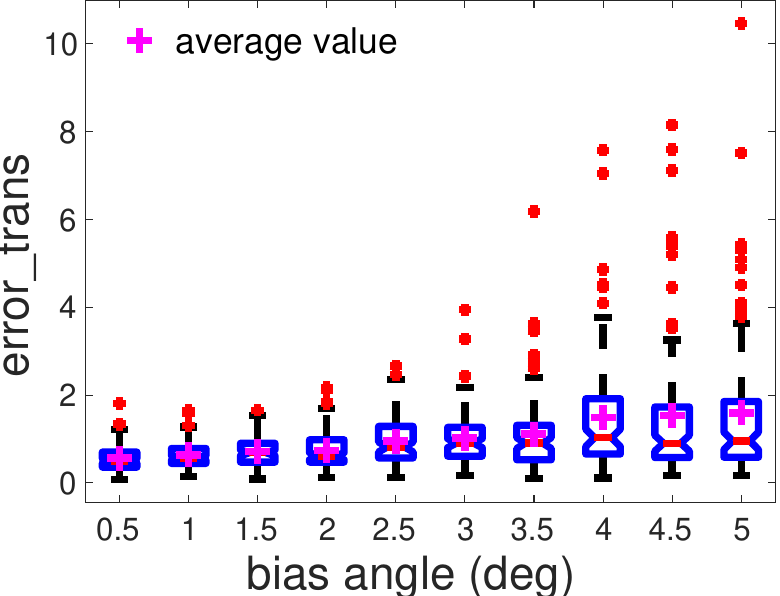}
		\label{fig:gbiasT}
	}
 \caption{\hl{The error distribution with various gravity bias. (a) Rotation error  (b) translation error.
 }}
 \label{fig:gbias}
 \centering
\end{figure}

\begin{figure}[!t]
    \centering
    \includegraphics[width=\linewidth]{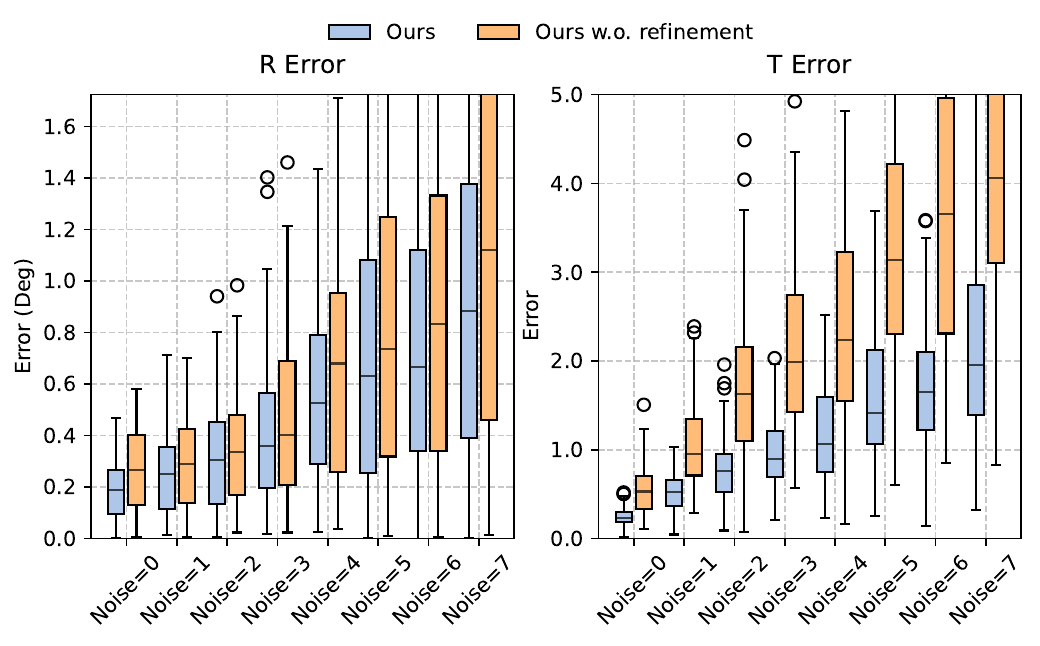}
 \caption{Performance comparison of the proposed method with or without pose refinement.}
 \label{fig:refinement_ablation}
 \centering
\end{figure}


\fakeparagraph{Gravity direction bias.}
\hl{It is also important to consider the vertical gravity direction bias, as IMU sensors may occasionally return values with deviations. To simulate such scenarios, we introduce a vertical direction bias with the parameters $N = 1000$, $\sigma = 1$, and $\omega = 0.5$. For the gravity bias, we randomly rotate the gravity vector in arbitrary directions, using varying rotation angles $\theta_g = \{0.5^\circ,1^\circ,\cdots,5^\circ\}$. Each trial is repeated 100 times, and the resulting error distributions are presented in Fig.~\ref{fig:gbias}.
The results indicate that our proposed method can still achieve acceptable absolute camera pose solutions, even with biased vertical directions. However, as the bias angle increases, both translation and rotation estimation errors grow accordingly, with a higher likelihood of large errors. This behavior is expected, as greater bias naturally leads to more significant deviations in pose estimation.
}

\fakeparagraph{Effect of pose refinement.}
To further validate the effectiveness of the proposed pose refinement module, we conduct ablation experiments comparing the method with and without pose refinement. The results, summarized in Fig.~\ref{fig:refinement_ablation}, demonstrate that incorporating the pose refinement module leads to improved performance.

\begin{figure}[!t]
	\subfigure[]{
		\includegraphics[width=0.45\linewidth]{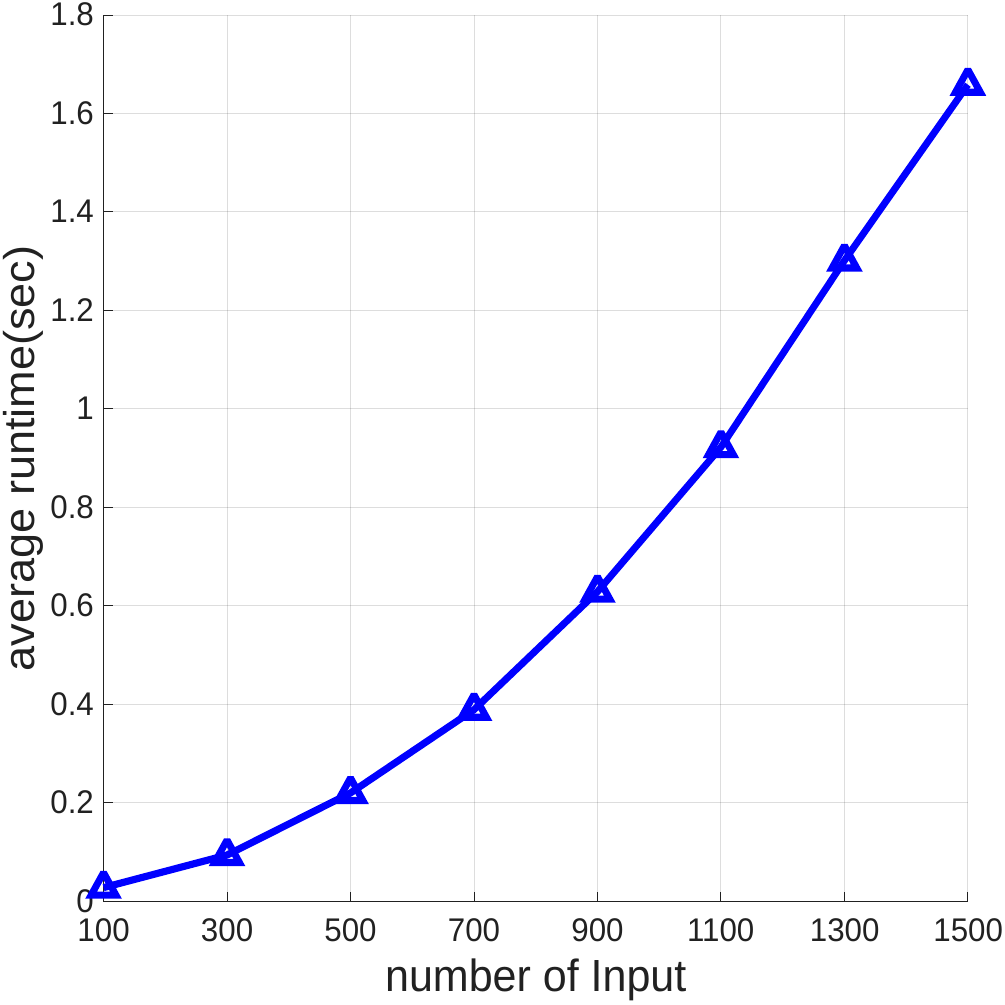}
		\label{fig:TimeN1}
	}
	\subfigure[]{
\includegraphics[width=0.45\linewidth]{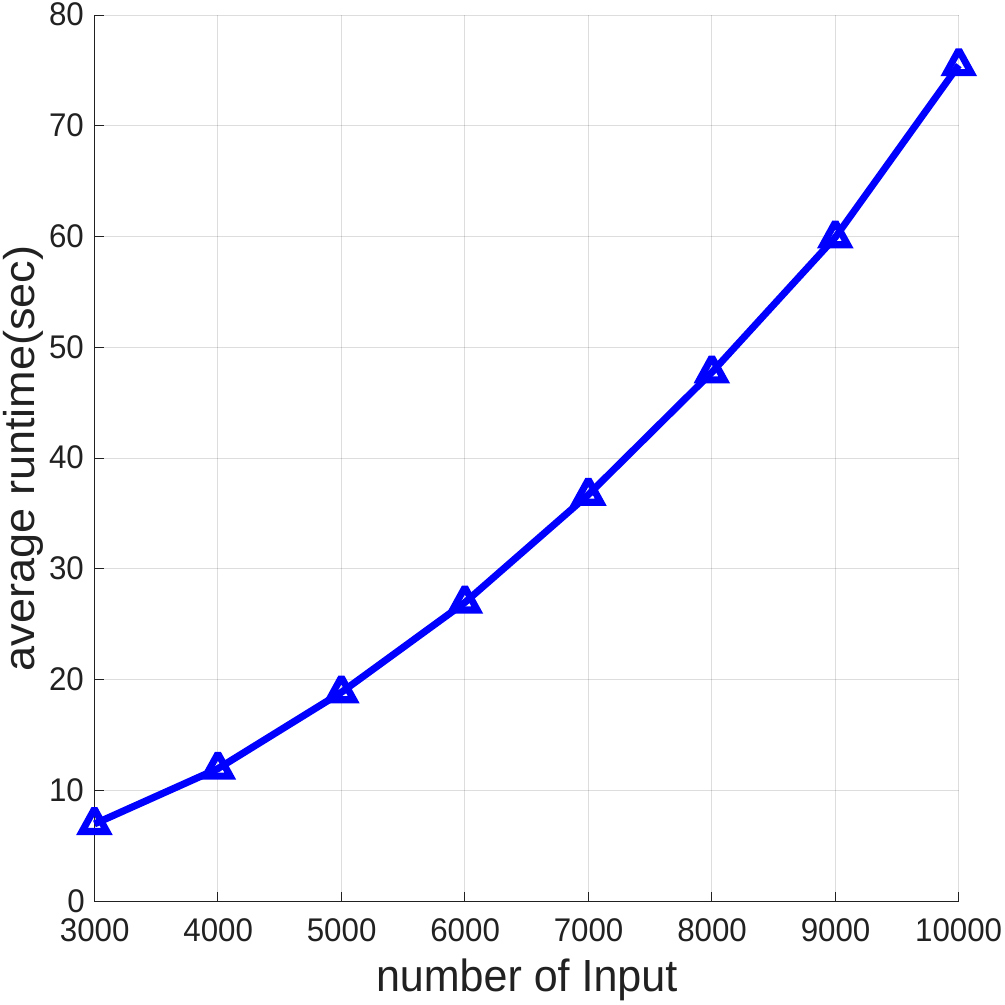}
		\label{fig:TimeN2}
	}
 \caption{Average runtime for different numbers of input pairs. (a) for $N \in [100, 1500]$, (b) for $N \in [3000, 10000]$.}
 \label{fig:Time_N}
 \centering
\end{figure}

\fakeparagraph{Time complexity analysis.}
In this part, we analyze the time complexity of our proposed method. As previously discussed, the runtime primarily consists of three components: $\mathbf{R}$ estimation, $\bm{t}$ estimation, and pose refinement. In Tab.~\ref{T:Time3part}, we present the average runtime for each component, along with the corresponding number of processed pairs, based on 100 experiments with the settings $N = 1000$, $\sigma = 1$, and $\omega = 0.5$. We also compare the runtime across different input sizes, specifically $N_1= \{100,300,\cdots,1500\}$ and $N_2 = \{3000,4000,\cdots,10000\}$. The average runtime results, computed over 100 trials, are illustrated in Fig.~\ref{fig:TimeN1} and Fig.~\ref{fig:TimeN2}.
The results indicate that $\mathbf{R}$ estimation is the most time-consuming step due to its need to process the largest number of pairs, $N$. Specifically, it requires $N(N-1)/2$ computations for each pair combination, resulting in a quadratic growth trend in runtime. Both curves in the figures clearly exhibit this trend, further confirming that $\mathbf{R}$ estimation dominates the overall time complexity. Notably, unlike noise-sensitive algorithms such as BnB, the runtime of our method remains unaffected by increases in noise level ($\sigma$) and outlier rate ($\omega$).

\begin{table}[t!]
\centering
\caption{Average runtime and number of input pairs for the three components.}
\begin{tabular}{c|c|c|c}
\toprule
Part & $\mathbf{R}$ estimation &$\bm{t}$ estimation & pose refinement \\
\midrule
Time(sec) & 0.70 & 0.21 & 0.004   \\
\midrule
N & 1000 & 100 & 91  \\
\bottomrule
\end{tabular}
\label{T:Time3part}
\centering
\end{table}

\begin{table}[t!]
\centering
\caption{Average runtime of our method under different parallelization settings.}
\begin{tabular}{c|c|c|c|c|c}
\toprule
Threads & 1 & 2 & 4&8&16\\
\midrule
Time(sec) & 8.05 & 4.09 & 2.15& 1.24 & 0.91     \\
\bottomrule
\end{tabular}
\label{T:TimeThread}
\centering
\end{table}

Furthermore, our method processes only two pairs at a time during $\mathbf{R}$ estimation, meaning that additional pairs can be handled concurrently. This characteristic makes our approach well-suited for parallel acceleration techniques, such as multithreading or GPU-based computation. In Tab.~\ref{T:TimeThread}, we compare the average runtime for 100 trials using varying numbers of threads, with $N = 1000$, $\sigma = 1$, and $\omega = 0.5$. The results show a significant reduction in runtime as the number of threads increases. Therefore, our method can be easily accelerated through parallel computing, making it feasible for real-time applications to some extent.

The above synthetic experimental results further validate the complementary advantages of our two-stage framework: the coarse estimation stage ensures stable and robust pose recovery under extreme outlier and noise conditions, while the refinement stage delivers consistent accuracy improvement on the basis of the coarse initial pose. The performance degradation of all compared baselines under high outlier rates also confirms the superiority of our proposed method in robust estimation for the gravity-constrained 1DoF rotation space.

\subsection{Real-World Data Experiments}
To evaluate the feasibility of our proposed method in real-world applications, we conduct experiments using real-world data. We utilize typical images and their corresponding 3D data (depth maps or 3D point sets) from three publicly available datasets: the TUM RGB-D dataset~\cite{sturm12iros}, the ETH3D dataset~\cite{Schops_2019_CVPR}, and the RobotCar dataset~\cite{doi:10.1177/0278364916679498}, all of which provide ground-truth trajectories.

\begin{figure}[t!]
    \centering
    \includegraphics[width=0.95\linewidth]{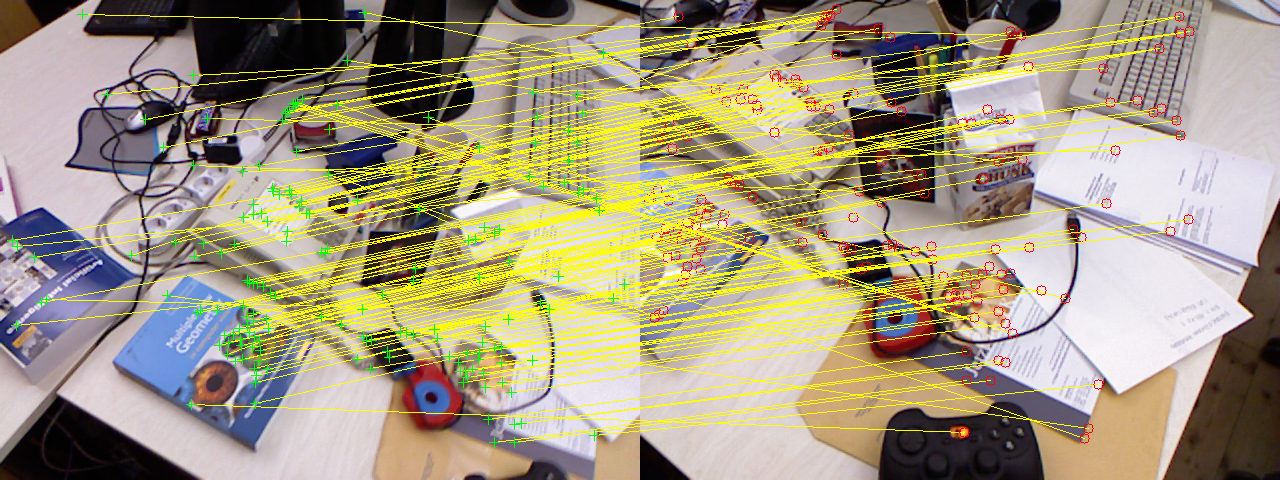}
    \caption{SIFT matching example from the freiburg1\_desk scene in the TUM RGB-D dataset.}
    \label{fig:SIFT_tum}
\end{figure}

\begin{figure}[t!]
    \centering
    \includegraphics[width=0.95\linewidth]{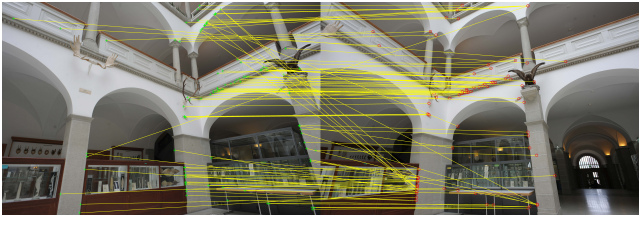}
    \caption{SIFT matching example from the exhibition scene in the ETH3D dataset. (For better visualization, the shown matches are downsampled).}
    \label{fig:SIFT_eth}
\end{figure}

\begin{figure}[t!]
    \centering
    \includegraphics[width=0.95\linewidth]{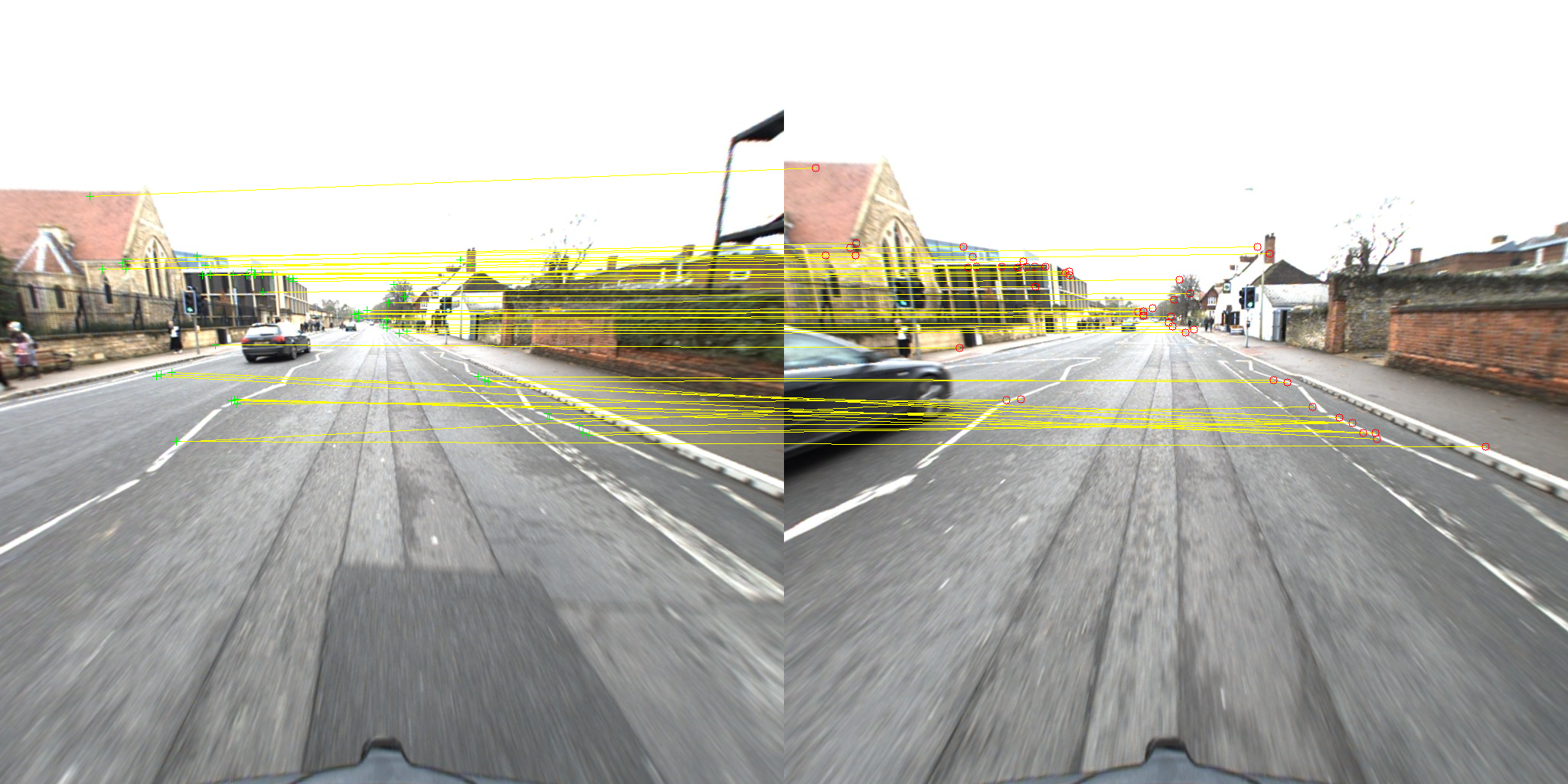}
    \caption{SIFT matching example from the overcast scene in the RobotCar dataset. (For better visualization, the shown matches are downsampled).}
    \label{fig:SIFT_robotcar}
\end{figure}

In our experiments, we select two images at a time to create 2D-3D pairs. For the first image, we obtain the bearing vectors $\bm{y}$ by using its 2D information, including pixel coordinates and the intrinsic matrix. For the second image, we acquire the corresponding 3D points $\bm{x}$ by using depth maps or Lidar point clouds. The 2D-3D pairs are generated by matching image pairs through feature descriptors, such as ORB~\cite{6126544} or SIFT~\cite{lowe2004distinctive}.
Since feature matching is not perfectly accurate, noise and outliers are inevitably introduced during the matching process. We then estimate the absolute camera pose for the first image and calculate the error relative to the ground-truth pose to assess the performance of our method.
To further highlight the robustness of our proposed method, we select non-adjacent frames for evaluation, as such pairs tend to produce more outliers during 2D-2D feature matching. Additionally, we intentionally avoid fine-tuning the feature-matching parameters to achieve ideal results. Examples from the three datasets are shown in Fig.~\ref{fig:SIFT_tum}, Fig.~\ref{fig:SIFT_eth}, and Fig.~\ref{fig:SIFT_robotcar}.

\begin{figure}[!t]
	\subfigure[]{
		\includegraphics[width=0.465\linewidth]{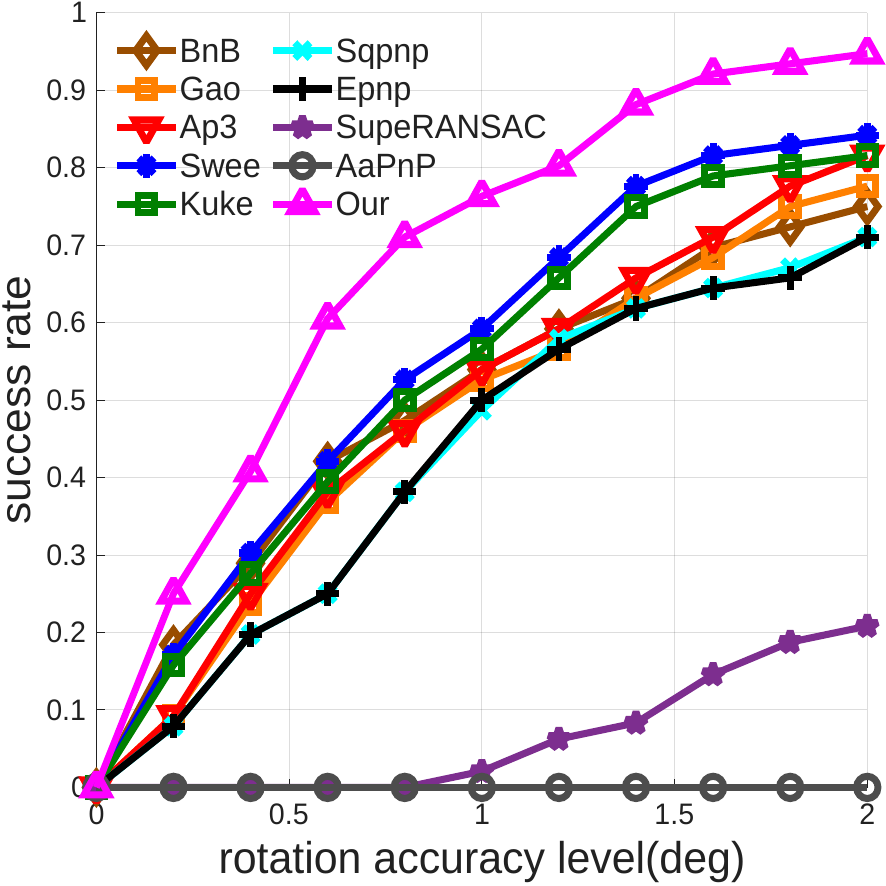}
		\label{fig:Tum_AcR}
	}
	\subfigure[]{
		\includegraphics[width=0.465\linewidth]{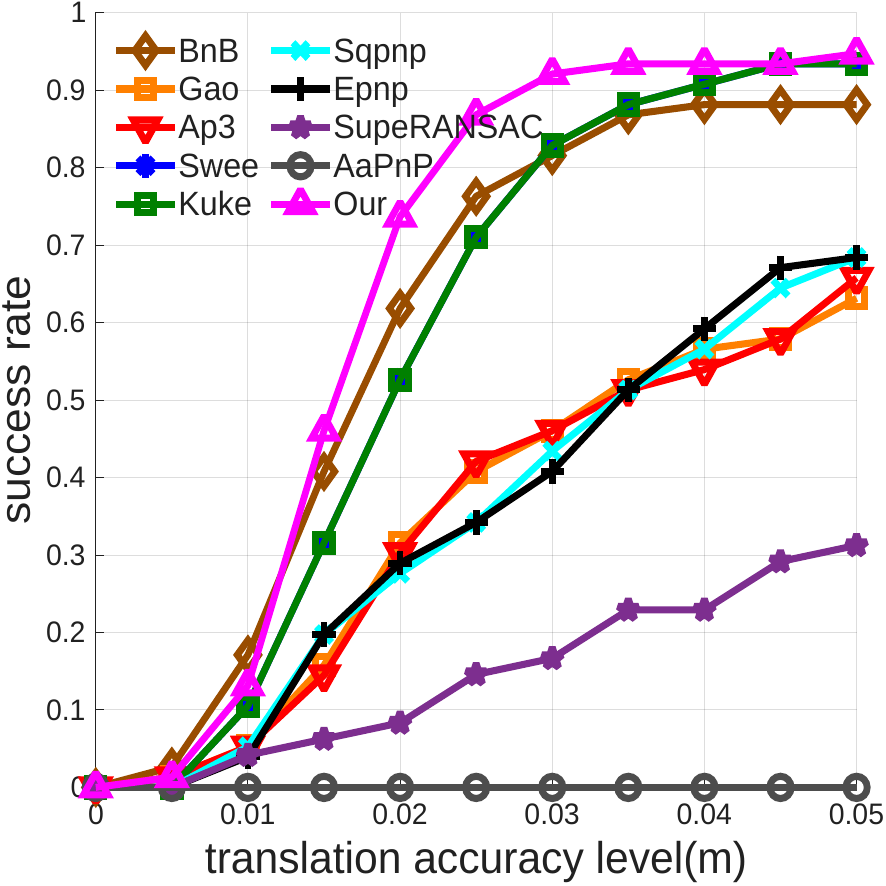}
		\label{fig:Tum_AcT}
	}
 \caption{\hl{Experimental results on selected images from the TUM RGB-D dataset (higher is better). 
 }}
 \label{fig:Tum_Ac}
 \centering
\end{figure}

\begin{figure}[!t]
	\subfigure[]{
		\includegraphics[width=0.465\linewidth]{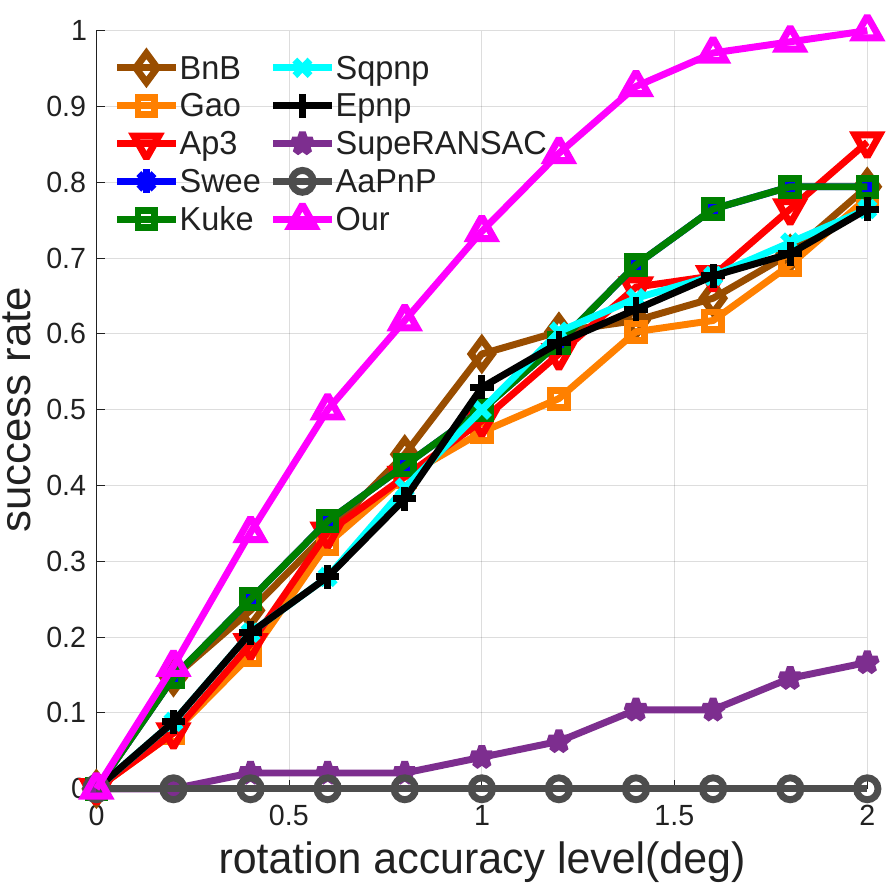}
		\label{fig:Eth_AcR}
	}
	\subfigure[]{
		\includegraphics[width=0.465\linewidth]{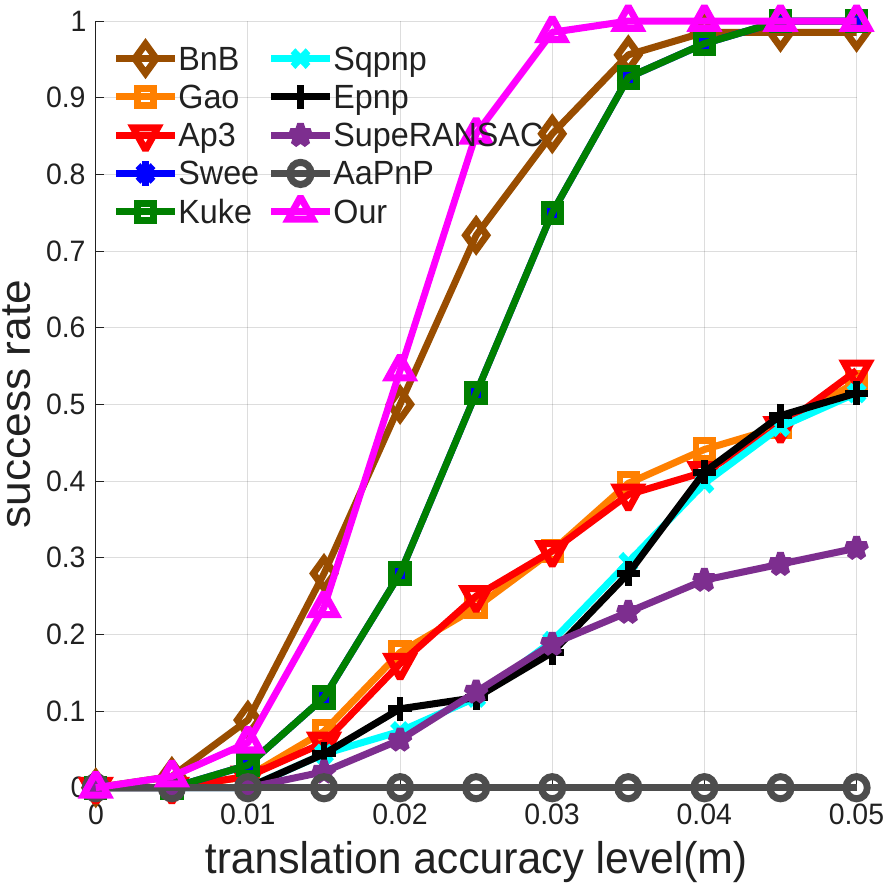}
		\label{fig:Eth_AcT}
	}
 \caption{\hl{Experimental results on selected images from the ETH3D dataset (higher is better). 
 }}
 \label{fig:Eth_Ac}
 \centering
\end{figure}

\begin{figure}[!t]
	\subfigure[]{
		\includegraphics[width=0.465\linewidth]{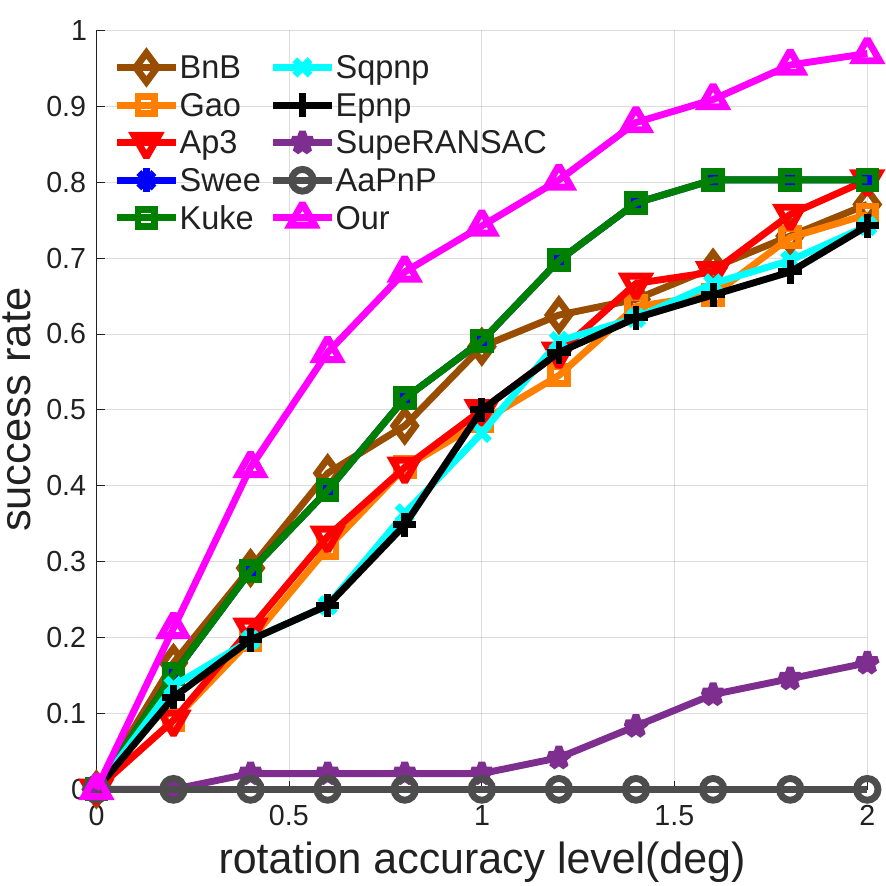}
		\label{fig:RobotCar_AcR}
	}
	\subfigure[]{
		\includegraphics[width=0.465\linewidth]{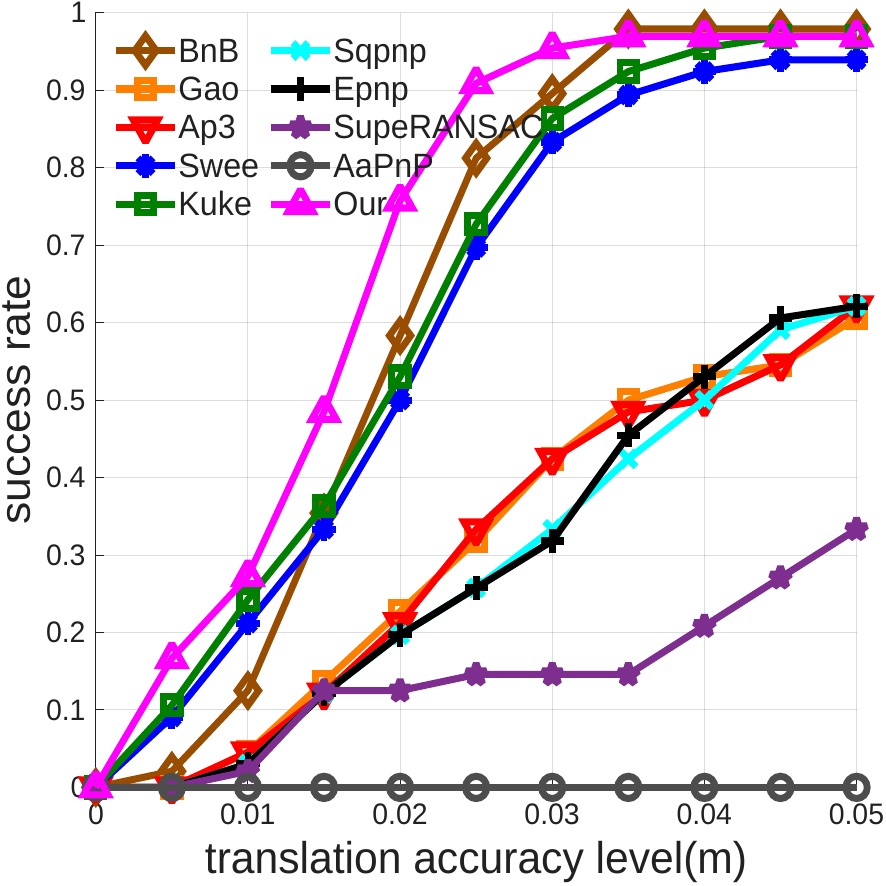}
		\label{fig:RobotCar_AcT}
	}
 \caption{\hl{Experimental results on selected images from the RobotCar dataset (higher is better). 
 }}
 \label{fig:RobotCar_Ac}
 \centering
\end{figure}

\begin{figure}[t!]
    \centering
    \includegraphics[width=\linewidth]{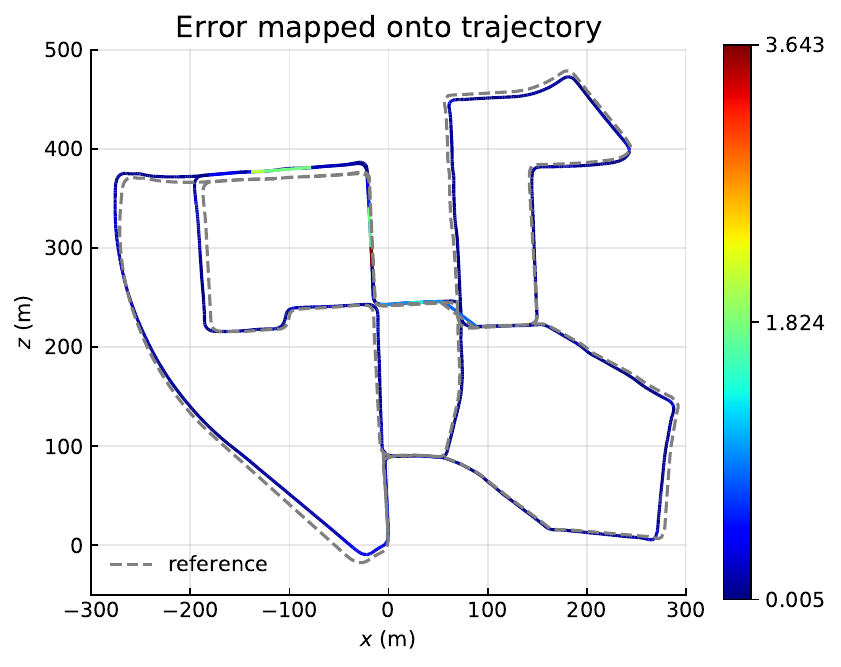}
    \caption{Error visualization of our method on a real-world trajectory \hl{from KITTI 01}. The estimated trajectory (solid line) is compared against the reference trajectory (dashed line), with colors indicating the magnitude of relative pose error (RPE). The predominantly blue coloring demonstrates our method's ability to maintain low errors across most of the trajectory, with only occasional higher errors (green/red) in limited segments.}
    \label{fig:error_map}
\end{figure}

When retrieving the 3D points corresponding to the matched pixels from the first image in the second image, we expand the search criteria to include pixels within a 10- to 20-pixel radius of the exact match to account for potential inaccuracies. For each dataset, we select approximately 100 image pairs of varying types and present the accuracy results in Fig.~\ref{fig:Tum_Ac}, Fig.~\ref{fig:Eth_Ac}, and Fig.~\ref{fig:RobotCar_Ac}. From the figures, we observe that our proposed method is effective in estimating the absolute camera pose when the vertical gravity direction is known. Additionally, it demonstrates greater robustness compared to many existing methods, particularly in scenarios with significant outlier contamination. However, a notable drawback of our approach is its longer runtime compared to RANSAC-type methods, especially when handling large input sizes ($N$). Fortunately, the method can be easily accelerated through parallel computing techniques, such as GPU processing or multithreading, making it theoretically viable for real-time applications. \hl{We also find that the compared method AaPnP~\cite{roch2024axes} performs significantly worse on the real-world datasets. This is mainly due to the fact that the real-world data contains mainly 6-DoF transformations, while the AaPnP~\cite{roch2024axes} is only focusing on 3-DoF transformations.}
Furthermore, we compare the time consumption of eight different methods in Tab.~\ref{T:RealDataTime}. Due to the variability in the number of pairs ($N$) generated from different image pairs, we report the average runtime by taking the middle 10\% of all outcomes for each dataset to provide a more representative comparison.

The real-world experimental results are consistent with the synthetic data, demonstrating that our method maintains strong robustness in complex real scenarios. Compared with the compared baselines, our method's performance advantage comes from the joint effect of the global voting module's outlier filtering capability and the refinement module's accuracy optimization, which makes it more suitable for practical applications.


\begin{table}[t!]
\centering
\caption{Runtime comparison of different methods on the real-world datasets.}
\resizebox{\linewidth}{!}
{\begin{tabular}{c|c|c|c|c|c|c|c|c|c|c}
\toprule
\diagbox{Dataset}{Time(sec)}{Method} & Swee & Kuke &Epnp&Gao& Sqpnp & Ap3 & BnB & SupeRANSAC & AaPnP & Our \\
\midrule
TUM RGBD & 0.53 & 0.54 & 0.03 &0.005&0.03 &0.01 &10 & \hl{0.06} & \hl{0.08} & 0.60 \\
\midrule
ETH3D  & 0.81 & 0.83 & 0.08 &0.04&0.08 &0.05 & 30 & \hl{0.07} & \hl{0.09} & 1.47 \\
\midrule
RobotCar& 0.66 & 0.67 & 0.06 &0.02&0.06 &0.04 &20 & \hl{0.05} & \hl{0.08} & 0.87 \\
\bottomrule
\end{tabular}}
\label{T:RealDataTime}
\centering
\end{table}

\begin{table}[t!]
\centering
\caption{\hl{Comparison between our method and the original ORB\_SLAM2\cite{murORB2} system. The evaluation metrics include Root Mean Square Error (RMSE), Sum of Squared Errors (SSE), minimum error (min), maximum error (max), and standard deviation (std). The metrics are computed based on the absolute pose error (APE) of every timestamp.}}
\begin{tabular}{c|c|c|c|c|c}
\toprule
Method & RMSE & SSE & min & max & std \\
\midrule
ORB\_SLAM2 & 16.40 & 361053.9 & \textbf{1.22} & 32.15 & 6.5687 \\
ORB\_SLAM2 + Ours & \textbf{9.36} & \textbf{115607.3} & 1.63 & \textbf{14.93} & \textbf{2.8524} \\
\bottomrule
\end{tabular}
\label{T:ORBSLAM-APE}
\centering
\end{table}

\begin{table}[t!]
\centering
\caption{\hl{Comparison between our method and the original ORB\_SLAM2\cite{murORB2} system. The evaluation metrics include Root Mean Square Error (RMSE), Sum of Squared Errors (SSE), minimum error (min), maximum error (max), and standard deviation (std). The metrics are computed based on the relative pose error (RPE) of every timestamp.}}
\begin{tabular}{c|c|c|c|c|c}
\toprule
Method & RMSE & SSE & min & max & std \\
\midrule
ORB\_SLAM2 & 1.029 & 1121.2 & 0.0101 & 31.36 & 0.9910 \\
ORB\_SLAM2 + Ours & \textbf{0.539} & \textbf{311.7} & \textbf{0.0058} & \textbf{16.70} & \textbf{0.5157} \\
\bottomrule
\end{tabular}
\label{T:ORBSLAM-RPE}
\centering
\end{table}

\begin{figure*}[!t]
    \centering
    \begin{minipage}{0.4\textwidth}
        \centering
        \includegraphics[width=\linewidth]{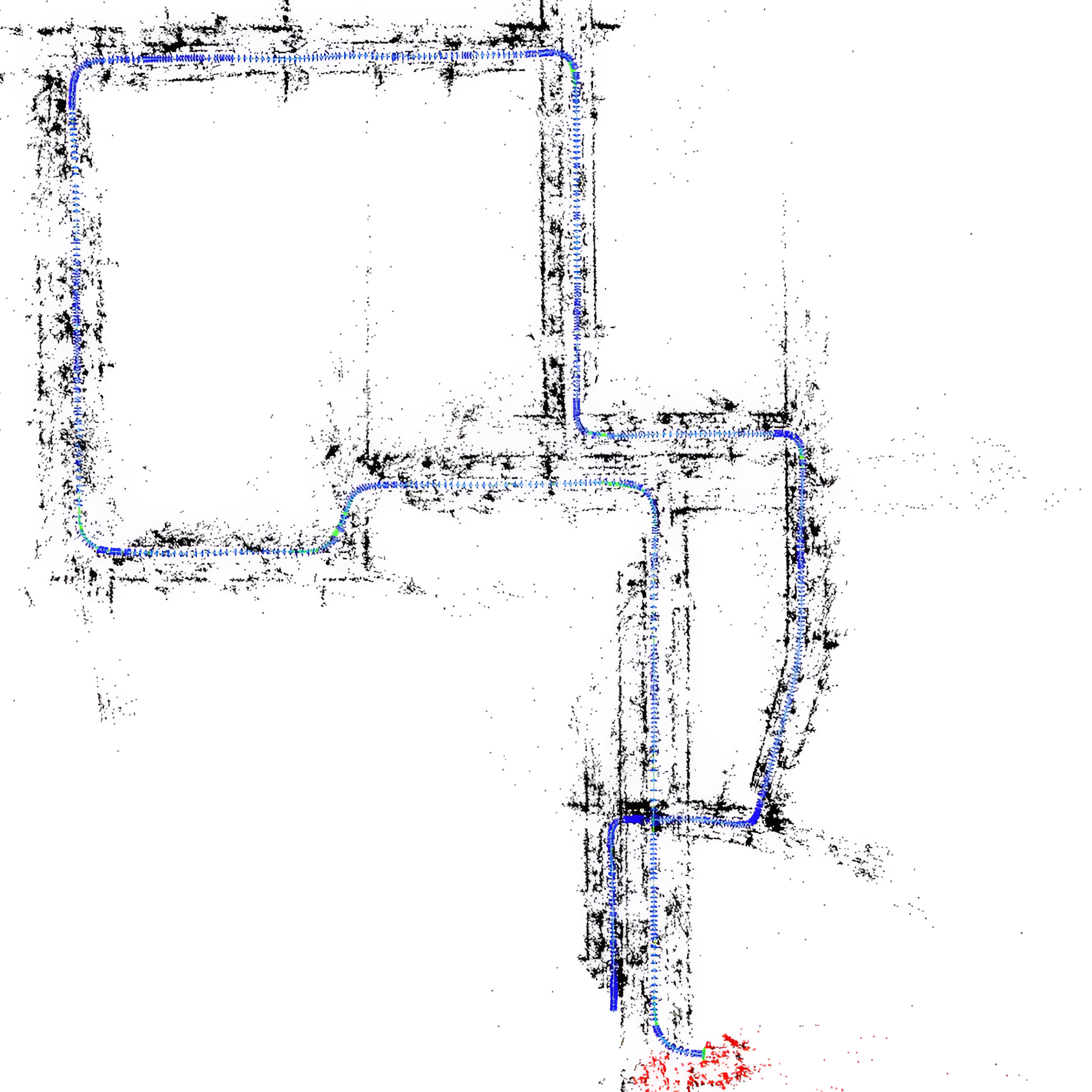}
        \caption*{(a)}
    \end{minipage}
    \begin{minipage}{0.4\textwidth}
        \centering
        \includegraphics[width=\linewidth]{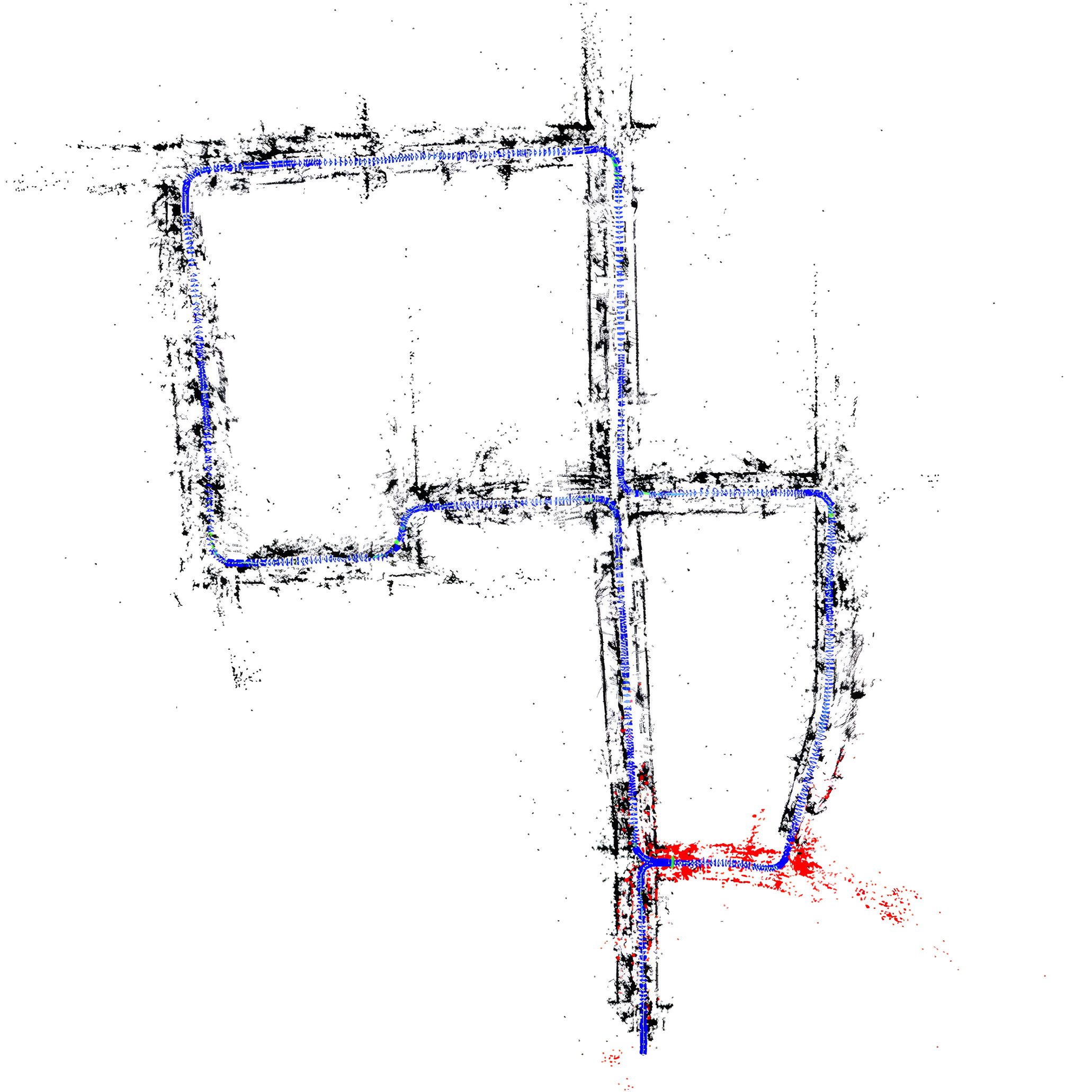}
        \caption*{(b)}
    \end{minipage}
    \\[1em] 
    \begin{minipage}{0.4\textwidth}
        \centering
        \includegraphics[width=\linewidth]{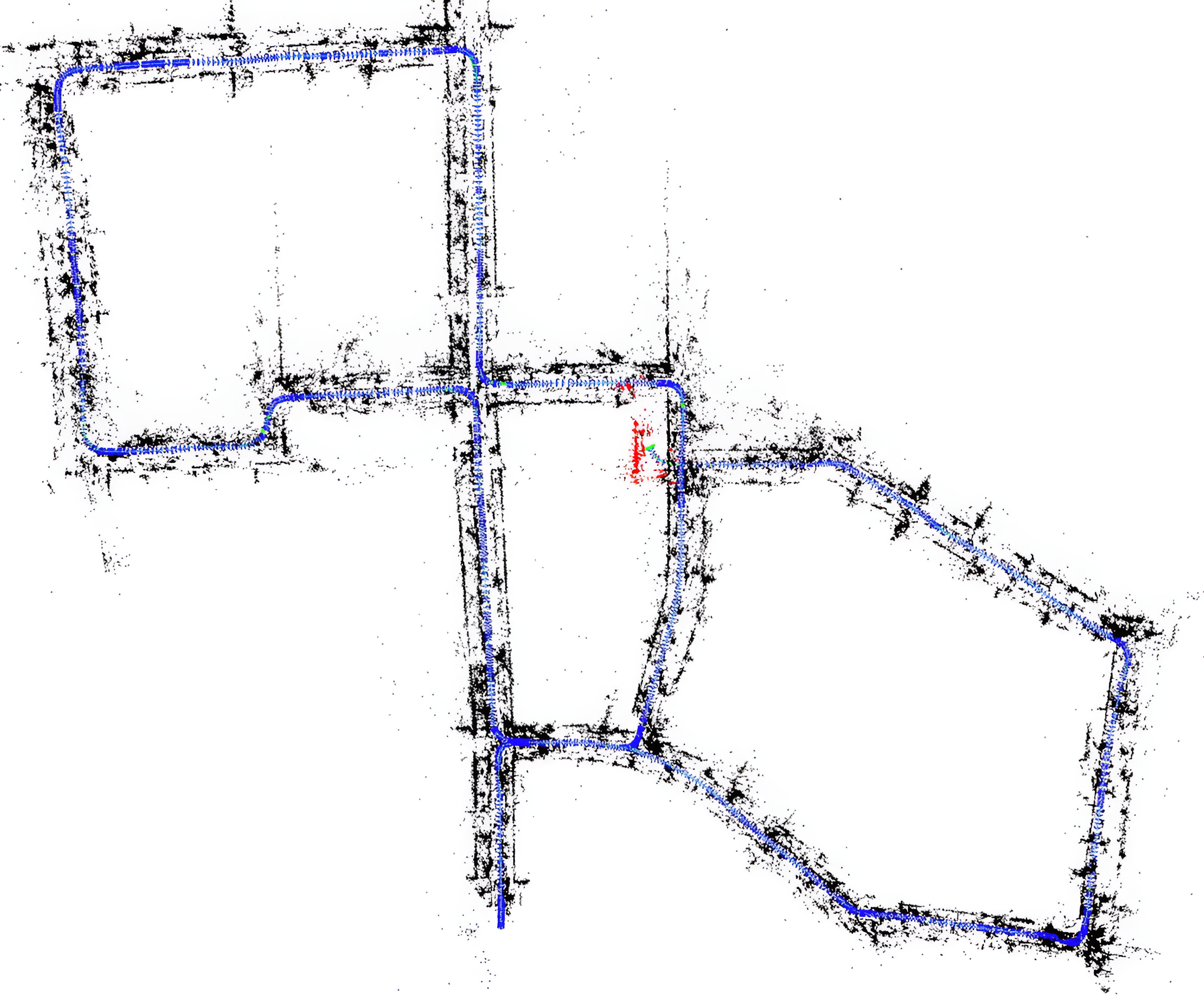}
        \caption*{(c)}
    \end{minipage}
    \begin{minipage}{0.4\textwidth}
        \centering
        \includegraphics[width=\linewidth]{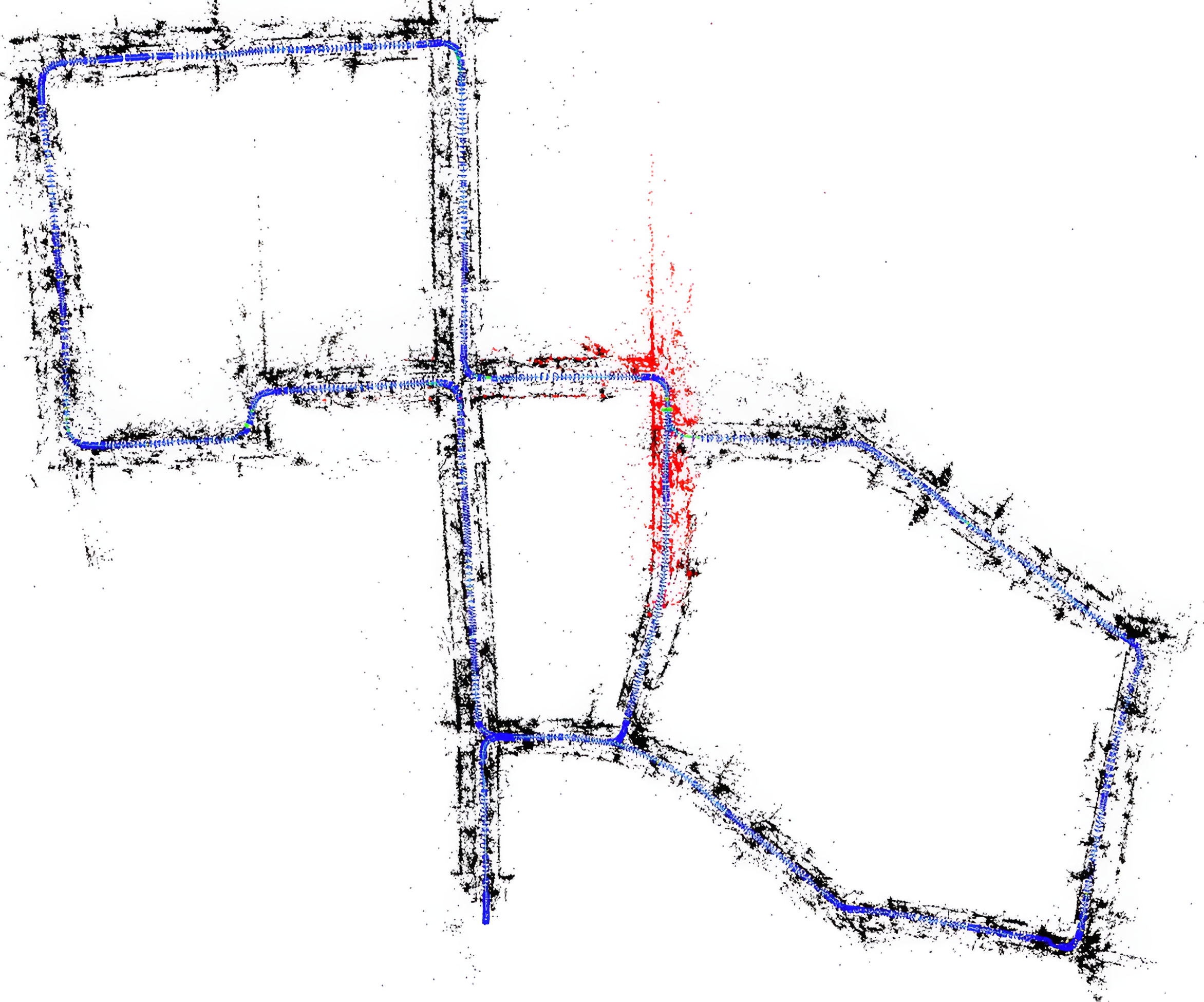}
        \caption*{(d)}
    \end{minipage}
    \caption{Visualization of our system's relocalization performance on Sequence 00 of the KITTI~\cite{geiger2013vision} dataset. (a) and (c) show the estimated trajectory (blue) before relocalization, where misalignment with the point cloud map (black) is evident due to accumulated drift. (b) and (d) illustrate the trajectory after relocalization, demonstrating that our gravity-prior-driven approach successfully corrects the alignment during the relocalization event. 
    }
    \label{fig:reloc_grid_1}
\end{figure*}


\subsection{Validation of ORB-SLAM2 integrated with our method on the KITTI dataset.}
To validate our method, we conducted comprehensive experiments by integrating our absolute pose estimation approach into ORB-SLAM2~\cite{murORB2}. Specifically, the original PnP solver of ORB-SLAM2 is replaced by our proposed solution. The experiments were conducted on the KITTI~\cite{geiger2013vision} dataset. \hl{We assessed the system's performance using Absolute Pose Error (APE) and Relative Pose Error (RPE) as the primary evaluation metric. APE and RPE were selected because they effectively measure the system's ability to maintain consistent pose estimation between consecutive frames while respecting the gravity constraint—a critical requirement for real-world applications where local accuracy and stability are essential.}
The experimental results confirm the effectiveness of our method. The visualization in Fig.~\ref{fig:error_map} shows that the estimated trajectory (solid blue line) closely aligns with the reference trajectory (dashed line) across a complex path. The color-coded error map indicates that our method maintains consistently low errors (predominantly blue regions) throughout most of the trajectory, with only minor deviations (green and red regions) observed in challenging segments. \hl{Quantitative comparisons in Tab.~\ref{T:ORBSLAM-APE} and Tab.~\ref{T:ORBSLAM-RPE} further validate these findings. For APE, our method achieves around 50\% less RMSE (9.36 vs. 16.40), SSE (115607.3 vs. 361053.9), max error (14.93 vs. 32.15), and std (2.8524 vs. 6.5687). For RPE, our method achieves an RMSE of 0.159, significantly lower than the original ORB-SLAM2's 0.176. Additionally, our approach yields a lower SSE (57.63 vs. 70.74), reduced maximum error (3.64 vs. 4.58), and a smaller standard deviation (0.1344 vs. 0.1550). The results from APE and RPE demonstrate the improved stability and reliability. These results indicate that our method not only enhances accuracy but also increases robustness in pose estimation.}

More graphically, as illustrated in Fig.~\ref{fig:reloc_grid_1}, our experimental results further demonstrate the effectiveness of our method through multiple relocalization test cases on the KITTI~\cite{geiger2013vision} dataset. Each test case presents paired visualizations showing system performance before and after relocalization events. The visualizations depict the estimated trajectories (shown in blue) alongside the constructed point cloud maps (in black), with current key points highlighted in red within the map structure. These results highlight the ability of our method to correct significant drift and improve trajectory alignment during relocalization.
In the pre-relocalization images, trajectory misalignments are consistently observed, where the estimated paths exhibit noticeable drift relative to the mapped environment. Such misalignments are common challenges in visual SLAM systems, particularly after prolonged operation or during challenging sequences. The post-relocalization images, however, clearly demonstrate our system's robust recovery capabilities across various scenarios. Following each relocalization event, our gravity-prior-driven approach successfully corrects trajectory alignments, effectively eliminating the previously accumulated drift. The corrected trajectories exhibit improved consistency with the surrounding environmental structure, validating our method's ability to ensure long-term reliability through effective relocalization.

The system-level experimental results fully validate the practical value of our proposed method. By replacing the original P$n$P solver of ORB-SLAM2 with our full two-stage framework, we achieve significant improvements in both trajectory accuracy and relocalization performance. This further confirms that our method's robust coarse estimation and high-precision refinement modules can work synergistically to address the practical challenges of drift accumulation and relocalization failure in real SLAM systems.


\section{Discussion and Analysis}
In this section, we systematically analyze the independent contribution of each core module and the synergistic effect of our two-stage cascaded framework, based entirely on the experimental results presented in previous sections. Our method is built on two tightly coupled stages with clear, complementary design goals: a coarse estimation stage for robust outlier filtering and reliable initial pose recovery, and a fine refinement stage for high-precision pose optimization. The contribution of each component is validated as follows.
\subsection{Contribution of the Coarse Estimation Stage}
The core of our coarse estimation stage is the proposed rotation-first decoupling strategy and 1D global voting module, which is the primary source of our method’s outlier robustness and computational efficiency. We conduct head-to-head comparisons with the SOTA gravity-aware 4DoF baseline~\cite{liu2023absolute}, which shares the identical problem setting and gravity prior with our work and differs only in its core robust estimation module and decoupling strategy. The consistent performance superiority of our method directly verifies the fundamental improvements brought by our proposed module, which overcomes the limitations of existing frameworks in both robustness and computational efficiency. We also compare our method with mainstream de facto standard robust P$n$P baselines, and the results further confirm that our voting scheme, which fully leverages the low-dimensional nature of the gravity-constrained rotation space, delivers stronger robustness than traditional random sampling-based robust estimation frameworks.
\subsection{Contribution of the Fine Refinement Stage}
The core of our fine refinement stage is the proposed non-minimal refinement algorithm tailored for the 4-DoFs gravity-constrained space, which is the primary source of the final pose accuracy improvement. We have completed dedicated module-level ablation experiments for this component, with a controlled variable setting that isolates the impact of the refinement module while keeping all other experimental conditions consistent. The ablation results clearly demonstrate that the proposed refinement module delivers a significant and consistent improvement in both rotation and translation accuracy, fully validating its independent contribution to the final performance, and filling the research gap of dedicated high-precision optimization for gravity-aware 4-DoFs pose estimation.

\subsection{Synergistic Effect of the Two-Stage Framework}
Our two-stage framework is designed with tight coupling and mutual reinforcement between the two stages. The coarse estimation stage filters the vast majority of outliers and provides a reliable initial pose, which ensures the stability and convergence of the subsequent refinement module. In turn, the fine refinement stage fully unlocks the accuracy potential of the high-quality inlier correspondences and initial pose provided by the coarse stage. The system-level integration experiments in the SLAM framework further validate this synergistic effect. The full two-stage pipeline delivers significant improvements in trajectory accuracy and relocalization performance, which cannot be achieved by either stage alone.

In summary, the existing complete set of experiments in our manuscript has rigorously validated the independent contribution of each core module, the superiority of our core design over strong baselines, and the rationality of our holistic cascaded framework, fully establishing the contribution closure of our work.
\section{Conclusion}
In this paper, we introduce a novel, efficient, and robust algorithm for absolute pose estimation. Our approach is based on a new transformation decoupling strategy that leverages geometric relations derived from the known gravity prior. By applying this strategy, the original 6-DoFs absolute pose estimation problem is reduced to a 4-DoFs problem: 1-DoF for the rotation angle and 3-DoFs for translation, significantly enhancing computational efficiency. To solve these subproblems, we employ a one-dimensional global voting algorithm for rotation estimation and use RANSAC to determine the translation. Additionally, we propose a novel pose refinement algorithm to further improve the accuracy of both rotation and translation estimates. Extensive experiments on synthetic data and real-world datasets (TUM RGB-D~\cite{sturm12iros}, ETH3D~\cite{Schops_2019_CVPR},  and RobotCar~\cite{doi:10.1177/0278364916679498}) demonstrate that our method outperforms existing SOTA approaches. To further validate the effectiveness of our method, we integrated it into ORB-SLAM2~\cite{murORB2} and conducted comprehensive experiments on the KITTI~\cite{geiger2013vision} dataset. The results demonstrate that our method effectively corrects substantial drift and significantly improves trajectory alignment during relocalization.

The module contribution analysis further confirms that the performance superiority of our method comes from the tight coupling of our three core components: the rotation-first decoupling strategy unlocks the low-dimensional advantage of the gravity-constrained space, the 1D global voting module ensures robust outlier filtering and initial pose recovery, and the dedicated refinement module delivers high-precision pose optimization. Our holistic framework addresses the key limitations of existing gravity-aware 4-DoFs pose solvers, and provides an efficient, robust, and high-precision solution for practical robotic and visual SLAM applications.

\section*{Acknowledgments}
This work is supported in part by Macao Science and Technology Development Fund (Grant No. 0022/2025/ITP1); and in part by the MANNHEIM-CeCaS (Central Car Server—Supercomputing for Automotive, No. 16ME0820).

\bibliographystyle{IEEEtran}
\bibliography{egbib}

\end{document}